\documentclass[10pt,twocolumn,letterpaper]{article}

\usepackage{cvpr}              

\usepackage{graphicx} 
\usepackage{caption} 

\usepackage{algorithm}
\usepackage{algpseudocode}
\usepackage{subcaption}
\usepackage{amsmath}
\usepackage{amssymb}
\usepackage{mathtools}
\usepackage{amsthm}
\usepackage{multirow}
\usepackage{xcolor}
\usepackage{colortbl}
\usepackage{hhline}
\usepackage{makecell}
\usepackage{array}
\usepackage{booktabs}

\theoremstyle{plain}
\newtheorem{thm}{Theorem}
\newtheorem{lem}[thm]{Lemma}

\theoremstyle{definition}

\newtheorem{assumption}{Assumption}

\DeclareMathOperator*{\argmin}{arg\,min}
\DeclareMathOperator*{\argmax}{arg\,max}

\definecolor{cvprblue}{rgb}{0.21,0.49,0.74}
\usepackage[pagebackref,breaklinks,colorlinks,allcolors=cvprblue]{hyperref}

\def\paperID{} 
\def\confName{CVPR}
\def\confYear{2026}

\title{Federated Learning with Feedback Alignment}

\author{Incheol Baek, Hyungbin Kim, Minseo Kim, Yon Dohn Chung \\
Korea University\\
{\tt\small \{inch307, hyungbinkim , minss126, ydchung\}@korea.ac.kr}
}

\begin{document}
\maketitle
\begin{abstract}
Federated Learning (FL) enables collaborative training across multiple clients while preserving data privacy, yet it struggles with data heterogeneity, where clients’ data are not distributed independently and identically (non-IID). This causes local drift, hindering global model convergence. To address this, we introduce \textbf{Federated Learning with Feedback Alignment (FLFA)}, a novel framework that integrates feedback alignment into FL. FLFA uses the global model’s weights as a shared feedback matrix during local training’s backward pass, aligning local updates with the global model efficiently. This approach mitigates local drift with minimal additional computational cost and no extra communication overhead. 

Our theoretical analysis supports FLFA’s design by showing how it alleviates local drift and demonstrates robust convergence for both local and global models. Empirical evaluations, including accuracy comparisons and measurements of local drift, further illustrate that FLFA can enhance other FL methods demonstrating its effectiveness.
\end{abstract}    
\section{Introduction}
\label{sec:intro}
Federated Learning (FL) has gained significant attention as an innovative approach enabling multiple clients to collaboratively train a shared model without centralizing their data~\cite{mcmahan2017communication}. In a typical FL setup, each client trains a local model using its own dataset and shares only the model updates with a central server. The server aggregates these updates to refine a global model, leveraging the diverse data from all clients.  However, FL faces a critical challenge, data heterogeneity~\cite{zhao2018federated, li2019convergence, sattler2019robust, mohri2019agnostic}. When clients' data are non-IID (not independently and identically distributed), local model updates deviate from the global optimum---a phenomenon called \textit{local drift}---degrading convergence and overall performance~\cite{wang2020tackling, karimireddy2020scaffold, acar2021federated, wang2020federated}.

Addressing data heterogeneity remains an active area of research. Recent studies have proposed regularization-based methods~\cite{FedProx,MOON,lee2024fedsol,fedacg} that add auxiliary terms to the local objective function to constrain local updates. Although these methods effectively reduce local drift, they introduce considerable computational overhead through additional loss terms and their associated gradient computations, which accumulate across multiple local epochs.

To address data heterogeneity without auxiliary loss terms, we take a fundamentally different approach: instead of constraining what the model learns, we modify how gradients are computed during training by revisiting the backpropagation (BP) algorithm itself. We draw inspiration from Feedback Alignment (FA)~\cite{lillicrap2016random}, originally proposed as a biologically plausible alternative to BP. In BP, weight matrix $W$ is used for both forward and backward passes. FA instead 
uses $W$ for forward but a fixed random matrix $B$ for backward passes. Despite this asymmetry, the network learns effectively as $W$ aligns with $B$.

Building on this principle, we propose \textbf{Federated Learning with Feedback Alignment (FLFA)}, which uses the global model weights as a shared feedback matrix $B$ for all clients during local training. Unlike traditional FA with random $B$, our $B$ equals the global weights received each round. Instead of each client using its own diverging local weights $W$ for BP, all clients use the same global model weights as their feedback matrix $B$. This shared feedback acts as a common reference that naturally aligns local updates across heterogeneous clients without auxiliary loss objective term.

Our main contributions are as follows:
\begin{itemize}
    \item \textbf{Theoretical foundation:} We develop a theoretical foundation that identifies the key factors contributing to local drift in federated learning and establishes how FLFA can mitigate this drift. We provide convergence analysis demonstrating that FLFA achieves robust convergence for both local and global models.

    \item \textbf{FLFA framework:} Deriving from the theoretical foundation, we introduce FLFA, a novel approach that employs the global model weights as a unified backward feedback mechanism. FLFA mitigates local drift while introducing only minimal computational overhead and no extra communication overhead. This makes FLFA highly compatible with existing FL algorithms and practical for real-world deployment.

    \item \textbf{Empirical validation:} Empirically, FLFA consistently improves accuracy when integrated into diverse FL methods and our ablation studies validate our design of FLFA, derived from the theoretical analysis. Through detailed local drift analysis, we provide empirical evidence that FLFA's shared feedback mechanism enables clients to implicitly perceive information about other clients through the global model.
\end{itemize}
\section{Related Work}

To mitigate the local drift in FL, recent studies have proposed adding an auxiliary term to the loss function or developing strategies to aggregate local models on the server. This section first provides an overview of existing FL algorithms that are related to our work.

\subsection{Mitigating local drift in federated learning}
FedAvg is the pioneering FL algorithm that aggregates locally trained models on the server using weighted averaging \cite{mcmahan2017communication}. Although FedAvg demonstrates strong performance in IID settings, it suffers from weight divergence under non-IID data distributions due to the independent optimization of local models.

To address this, several studies focus on regularizing the local training process by adding an auxiliary term to the local objective function. Model-contrastive federated learning (MOON) \cite{MOON} improves FL robustness by employing contrastive learning between local and global model representations. By enforcing consistency across client updates, MOON alleviates representation drift. FedProx \cite{FedProx} introduces a proximal term that constrains local updates, preventing them from straying too far from the global model.

Other recent approaches tackle label distribution skew by intervening at the classifier's output. FedRS \cite{li2021fedrs} modifies the final layer by restricting the softmax function to locally available classes. Similarly, FedLC \cite{fedlc} regularizes the model's logits before the softmax calculation to prevent over-confidence in local majority classes. 

Although these methods effectively mitigate local drift, they rely on adding auxiliary terms or modifications to the local objective function, which can introduce additional computational costs. In contrast, our approach modifies the backpropagation process itself without adding any new terms to the loss function. This results in minimal additional cost and ensures high compatibility, allowing the method to integrate well with other FL methods.

\subsection{Learning with feedback alignment}
Feedback Alignment (FA) \cite{lillicrap2016random} is a biologically plausible learning algorithm that addresses the biological impracticality of backpropagation, which constructs a backpropagation path that is symmetric to the forward path. FA shows that the network is effectively trained even if a fixed random matrix is used instead of a transpose of the forward weights during backpropagation. This is possible because the forward weights are gradually aligned in the direction of the feedback signal as the learning progresses.

There are studies that apply Direct Feedback Alignment (DFA) \cite{nokland2016direct} to improve the computational efficiency of federated learning \cite{colombo2023tifed, jung2023lafd}. However, DFA is conceptually distinct from FA. DFA propagates the error directly and independently from the output layer to each hidden layer, allowing each layer to be trained in parallel. The aim is to reduce computational cost by completely omitting backpropagation computation. On the other hand, our goal is not to omit backpropagation computations, but to match the gradient between clients through a shared feedback path to mitigate local drift and improve the performance of the model.

\textbf{Our position.} Our approach, FLFA, aligns local updates by using global model weights as a unified backward feedback pathway during backpropagation. Unlike regularization methods that add auxiliary terms, FLFA modifies the gradient computation itself, using these shared global weights as a feedback matrix to align each client's updates toward the global objective. This novel application of FA's alignment property is the key contribution of our work and fundamentally distinguishes our approach from existing DFA-based studies focused on computational efficiency. To the best of our knowledge, we are the first to apply FA to address local drift in FL.
\section{Preliminaries}
\subsection{Federated learning}
FL is a decentralized framework where multiple clients collaboratively train 
a shared model without centralizing their raw data. Suppose there are $N$ 
clients, each with local dataset $\mathcal{D}_i=\{(x_{ij},y_{ij})\}_{j=1}^{N}$. 
The objective is:
\begin{equation}
    \underset{w}{\mathrm{arg\ min}}\ \mathcal{J}(w)=\sum_{i=1}^N\cfrac{|\mathcal{D}_i|}{|\mathcal{D}|}J_i(w),
\end{equation}
where $J_i(w)=\mathbb{E}_{(x,y) \sim \mathcal{D}_i} [ \ell (w;x,y)]$ is the 
local objective, $\ell (w;x,y)$ is the loss, and $|\mathcal{D}|=\sum_{i=1}^N|\mathcal{D}_i|$.

\subsection{Feedback alignment}
BP requires weight transposes $(W^T)$ for backward passes, which is biologically 
implausible~\cite{lillicrap2016random}. FA instead uses a fixed random feedback 
matrix $B$. For a network at training step $k$, layer $l$'s pre-activation is:
\begin{equation}
    z_l^k = w_l^k h_l^k,
\end{equation}
where $h_l^k = f( z_{l-1}^k)$. After $s$ steps with learning rate $\eta$:
\begin{equation}
    w_l^s = w_l^0 - \eta \sum_{k=0}^{s-1} \delta_l^k (h_l^k)^T,
\end{equation}
In BP, the error signal $\delta_l^k$ propagates as:
\begin{equation}
    \delta_l^k = \big((w_{l+1}^k)^T \delta_{l+1}^k\big) \odot f'(z_l^k),
\end{equation}
where $\odot$ is element-wise multiplication. FA replaces $W$ with fixed $B$:
\begin{equation}
    \delta_l^k = \big((B_{l+1}^k)^T \delta_{l+1}^k\big) \odot f'(z_l^k).
\end{equation}
Despite random $B$, forward weights $W$ gradually align with the feedback 
signal during training~\cite{lillicrap2016random}.

\section{Proposed Method} \label{sec:method}
In this section, we present the theoretical foundation of FLFA and explain how it mitigates local drift in FL caused by data heterogeneity.

\subsection{Theoretical foundation} \label{sec:theoretical_foundation}
To understand how FLFA addresses the challenges of non-IID data, we first analyze local drift in a standard FL setting. Consider a scenario with clients $i$ and $j$, focusing on the $l$-th layer of a neural network at communication round $r$ and local training step $k$. We denote the weight matrix for client $i$ at layer $l$, round $r$, and step $k$ as $w_{i,l}^{(r,k)}$.

\textbf{Setup.} At the start of round $r$, all clients receive the same global model:
\begin{equation}
    w_{i,l}^{(r,0)} = w_{j,l}^{(r,0)} = W_{l}^{r}
\end{equation}
where $W_{l}^{r}$ is the global model's weight matrix for layer $l$ at round $r$. After $s$ local training steps with learning rate $\eta$, client $i$'s weight becomes:
\begin{equation}
    w_{i,l}^{(r,s)} = w_{i,l}^{(r,0)} - \eta \sum_{k=0}^{s-1} \delta_{i,l}^{(r,k)} (h_{i,l}^{(r,k)})^T
\end{equation}
where $h_{i,l}^{(r,k)}$ is the layer input and $\delta_{i,l}^{(r,k)}$ is the error signal propagated via backpropagation:
\begin{equation}
    \delta_{i,l}^{(r,k)} = (w_{i,l+1}^{(r,k)})^T \delta_{i,l+1}^{(r,k)} \odot f'(z_{i,l}^{(r,k)})
\end{equation}
Note that in standard BP, the error signal uses the \textit{local weight transpose} $(w_{i,l+1}^{(r,k)})^T$, which diverges across clients as training progresses.

\textbf{Quantifying local drift.} We measure local drift as the difference in weight updates between two clients. Let $\Delta^r_{i, l} = w_{i,l}^{(r,s)} - w_{i,l}^{(r,0)}$ denote client $i$'s update. The drift between clients $i$ and $j$ is:
\begin{equation}
    \Delta^r_{i, l} - \Delta^r_{j, l} = -\eta \sum_{k=0}^{s-1} [\delta_{i,l}^{(r,k)} (h_{i,l}^{(r,k)})^T - \delta_{j,l}^{(r,k)} (h_{j,l}^{(r,k)})^T  ]
\end{equation}
The larger this difference, the more clients learn in divergent directions, impeding global model convergence. To quantify this, we assume bounded inputs and error signals: $\Vert h_{i,l+1}^{(r,k)} \Vert \leq \Tilde{x}$ and $\Vert \delta_{i,l+1}^{(r,k)} \Vert \leq \Tilde{\delta}$. Under these assumptions, we derive (see Section \ref{sec:supp:derivation} in supplementary for full derivation):
\begin{align} \label{eq:delta_diff}
&\Vert \Delta^r_{i, l} - \Delta^r_{j, l} \Vert_F \nonumber \\
&\leq \eta \sum_{k=0}^{s-1} \big[ \Tilde{x} \Vert w_{i,l+1}^{(r,k)}\Vert \cdot \Vert\delta_{i,l+1}^{(r,k)} - \delta_{j,l+1}^{(r,k)}\Vert \\
& + \Tilde{x}\Tilde{\delta} \Vert w_{i,l+1}^{(r,k)} - w_{j,l+1}^{(r,k)}\Vert +  \Tilde{\delta}\Vert w_{j,l+1}^{(r,k)}\Vert \cdot \Vert h_{i,l}^{(r,k)} - h_{j,l}^{(r,k)}  \Vert]\nonumber
\end{align}

\textbf{Key observation.} The bound depends on three factors: (1) error signal divergence $\Vert\delta_{i,l+1}^{(r,k)} - \delta_{j,l+1}^{(r,k)}\Vert$, (2) weight divergence $\Vert w_{i,l+1}^{(r,k)} - w_{j,l+1}^{(r,k)}\Vert$, and (3) input divergence $\Vert h_{i,l}^{(r,k)} - h_{j,l}^{(r,k)}\Vert$. This bound indicates that minimizing differences in inputs ($h$), error signals ($\delta$), and weights ($w$) between clients can reduce local drift.

\begin{figure*}[!ht]
    \centering
    \includegraphics[width=0.95\linewidth]{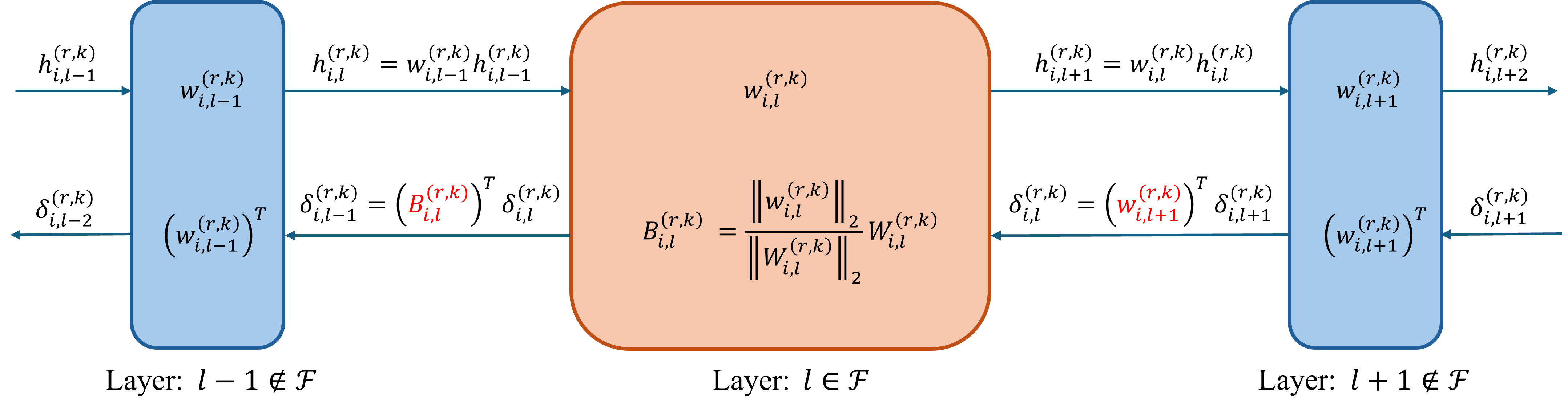}
    \caption{Illustration of FLFA's selective layer application mechanism. In the backward pass (bottom arrows), the middle layer (orange) uses FLFA: it computes the error signal for the previous layer using the feedback matrix $B_{i,l}^{(r,k)}$, which is derived from the global model weights $W_{i}^r$. The forward pass (top arrows) and other layers in the backward pass follow standard BP.}
    \label{fig:flfa}
\end{figure*}

\textbf{FLFA's approach.} Instead of using diverging local weights $(w_{i,l+1}^{(r,k)})^T$ for backpropagation, FLFA uses a shared feedback matrix $B$ derived from the global model. Specifically, in FLFA, the error signal becomes:
\begin{equation}
    \delta_{i,l}^{(r,k)} = \big((B_{i,l+1}^{(r,k)})^T \delta_{i,l+1}^{(r,k)} \big) \odot f'(z_{i,l}^{(r,k)})
\end{equation}
We set $B_{i,l+1}^{(r,k)} = B_{j,l+1}^{(r,k)} = W_{l+1}^r$ for all clients and steps within round $r$. This yields a tighter drift bound:
\begin{equation} \label{eq:fa_delta_diff}
    \begin{split}
&\Vert \Delta^r_{i, l} - \Delta^r_{j, l} \Vert_F \\
&\leq \eta \sum_{k=0}^{s-1} \big[ \Tilde{x} \Vert B_{i,l+1}^{(r,k)}\Vert \cdot \Vert\delta_{i,l+1}^{(r,k)} - \delta_{j,l+1}^{(r,k)}\Vert \\
& + \Tilde{\delta}\Vert B_{j,l+1}^{(r,k)}\Vert \cdot \Vert h_{i,l}^{(r,k)} - h_{j,l}^{(r,k)}  \Vert]
    \end{split}
\end{equation}

\textbf{Result.} The weight divergence term $\Tilde{x}\Tilde{\delta} \Vert w_{i,l+1}^{(r,k)} - w_{j,l+1}^{(r,k)}\Vert$ is completely eliminated. Since all clients use the same $B = W^r$, this term becomes zero. Furthermore, the shared feedback signal implicitly reduces error signal divergence $\Vert\delta_{i,l+1}^{(r,k)} - \delta_{j,l+1}^{(r,k)}\Vert$ by anchoring all clients' backward passes to the global model.

\textbf{Adaptive scaling.} While Eq.~\eqref{eq:fa_delta_diff} eliminates weight divergence, $B$ remains fixed while local weights $w$ evolve during training. If $\Vert w^{(r,k)}\Vert$ decreases significantly, the fixed $\Vert B\Vert$ may cause the FA bound to exceed the BP bound. To prevent this, we adaptively scale $B$ to match local weight norms: $\Vert B_{i,l+1}^{(r,k)}\Vert = \Vert w_{i,l+1}^{(r,k)}\Vert$. With a factor $\alpha = \frac{\Vert w_{j,l+1}^{(r,k)}\Vert}{\Vert w_{i,l+1}^{(r,k)}\Vert}$, the bound becomes:
\begin{align}\label{eq:fa_delta_diff_2}
&\Vert \Delta^r_{i, l} - \Delta^r_{j, l} \Vert_F \nonumber \\
&\leq \eta \sum_{k=0}^{s-1} \big[ \Tilde{x} \Vert w_{i,l+1}^{(r,k)}\Vert \cdot \Vert\delta_{i,l+1}^{(r,k)} - \delta_{j,l+1}^{(r,k)}\Vert \\
& + \Tilde{x}\Tilde{\delta} |1-\alpha| \cdot \Vert B_{i,l+1}^{(r,k)} \Vert +  \Tilde{\delta}\Vert w_{j,l+1}^{(r,k)}\Vert \cdot \Vert h_{i,l}^{(r,k)} - h_{j,l}^{(r,k)}  \Vert]\nonumber
\end{align}

\textbf{Why this is tighter than BP:} If client weight norms remain similar ($\Vert w_{i,l+1}^{(r,k)} \Vert \approx \Vert w_{j,l+1}^{(r,k)}\Vert$, i.e., $\alpha \approx 1$), then $|1-\alpha| \approx 0$. This is a \textit{much weaker condition} than requiring $\Vert w_{i,l+1}^{(r,k)} - w_{j,l+1}^{(r,k)}\Vert \approx 0$ (identical weights). In practice, even when weights diverge significantly in direction, their norms often remain comparable. Thus, FLFA achieves a tighter drift bound than BP while using only simple norm adjustments.

\textbf{On using global weights as $B$}: While any fixed matrix $B$ could serve as feedback, using $B = W^r$ offers three advantages. \textbf{(1)} Zero communication overhead: the global model is already transmitted each round. In contrast, a random $B$ would require extra communication \textbf{(2)} Tighter convergence bounds: our analysis (Section~\ref{sec:convergence}) shows that $B = W^r$ minimizes the approximation error term in convergence guarantees. \textbf{(3)} Semantic alignment: the global model represents aggregated knowledge from all clients, making it a natural reference point for coordinating local updates.

\subsection{FLFA framework}
Based on the theoretical foundation, we propose the framework of FLFA. The FLFA algorithm is designed to integrate feedback alignment principles into FL, mitigating local drift across distributed clients. FLFA is illustrated in \Cref{fig:flfa} and the algorithm of FLFA is presented in Section \ref{sec:supp:alg} of the supplementary materials. Given the total number of communication rounds ($R$), the number of local epochs ($E$), the learning rate ($\eta$), and a set of layers using FA ($\mathcal{F}$), the FLFA framework operates as follows:
\begin{itemize}
    \item \textbf{Initialization:} The server initializes the global model at round $r = 0$, denoted $\mathbf{W}^r$, which consists of weights for $L$ layers: $\mathbf{W}^r = \{ W_1^r, W_2^r, \ldots, W_L^r \}$.
    \item \textbf{Client selection and initialization:} In each communication round $r$, a subset of clients $\mathcal{U}^r$ is selected. Each client $i \in \mathcal{U}^r$ initializes its local model $\mathbf{w}_i^r$ with the global model, where $\mathbf{w}_i^r = \{ w_{i,1}^r, w_{i,2}^r, \ldots, w_{i,L}^r \}$. For layers configured to use FA (denoted by a set $\mathcal{F}$), feedback matrices $\mathbf{B}_i^r = \{ B_{i, l}^r \mid l \in \mathcal{F} \}$ are initialized such that $B_{i,l}^r \gets W_l^r$. Layers not in $\mathcal{F}$ use standard BP.
    \item \textbf{Local model updates:} Each client performs $E$ epochs on its local dataset $\mathcal{D}_i$. During training, layers in $\mathcal{F}$ use FLFA, while other layers use standard BP:
    \begin{itemize}
        \item For layers $l \in \mathcal{F}$ (using FA), the error signal $\delta_{i,l-1}$ in the backward pass is computed using the feedback matrix $B_{i,l}^r$ instead of the local weight $w_{i,l}^r$.
        \item For layers not in $\mathcal{F}$ (using BP), the standard BP error signal is computed.
    \end{itemize}
    \item \textbf{Adaptive feedback matrix:} After each batch update, the feedback matrices are scaled to match the norm of the updated local weights. This follows the adaptive scaling derived in Eq.\eqref{eq:fa_delta_diff_2}.
    \item \textbf{Global aggregation:} After local updates, the server aggregates the local models $\mathbf{w}_i^r$ from all clients in $\mathcal{U}^r$ into a new global model $\mathbf{W}^{r+1}$, using a weighted average based on the size of each client’s dataset.
\end{itemize}

\subsection{Discussion} \label{sec:discussion}
\textbf{Selective application of FA.} FLFA need not be applied to all layers. Applying it to even a single layer is enough to reduce drift throughout the network via a cascading effect: since error signals $\delta$ propagate backward layer-by-layer, alignment at one layer reduces divergence in error signals and inputs at preceding layers (as shown in Eq.~\eqref{eq:delta_diff}). 

In practice, we select \textbf{one layer} per round based on gradient alignment between clients. Let $\mathbf{g}_i^l$ denote the flattened gradient update for client $i$ at layer $l$, and $\bar{\mathbf{g}}^l = \frac{1}{K}\sum_{i=1}^K \mathbf{g}_i^l$ the mean update. We compute:
\begin{equation} \label{eq:cos}
z^l = \frac{1}{K}\sum_{i=1}^K \frac{\mathbf{g}_i^l \cdot \bar{\mathbf{g}}^l}{\|\mathbf{g}_i^l\| \|\bar{\mathbf{g}}^l\|}
\end{equation}
where higher $z^l$ indicates greater alignment. We select either $\argmin_l z^l$ (most divergent layer) to directly correct drift, or $\argmax_l z^l$ (most aligned layer) to reinforce stability across the network.

\textbf{Minimal overhead.} On the client side, FLFA only adds norm adjustment of the feedback matrix (computationally negligible) and requires zero extra communication since $B$ is derived from already-transmitted global weights. On the server side, layer selection involves light cosine similarity computations on already-aggregated gradients.

\textbf{Intuition for FLFA.} FLFA creates an asymmetry between forward and backward passes: in the forward pass, local weights $w_i^r$ extract client-specific features; in the backward pass, the shared feedback matrix $B = W^r$ propagates error signals 
anchored to the global model. Intuitively, all clients see the same backward gradient structure through $B$, enabling implicit coordination despite heterogeneous data---without any explicit communication of client information.

\begin{table*}[ht!]
\centering
\small
\begin{tabular}{l cc cc cc cc cc cc}
\toprule
\multirow{2}{*}{Method} & \multicolumn{2}{c}{BloodMNIST} & \multicolumn{2}{c}{OrganCMNIST} & \multicolumn{2}{c}{OrganSMNIST} & \multicolumn{2}{c}{PathMNIST} & \multicolumn{2}{c}{FMNIST} & \multicolumn{2}{c}{CIFAR-10} \\
\cmidrule(lr){2-3}\cmidrule(lr){4-5}\cmidrule(lr){6-7}\cmidrule(lr){8-9}\cmidrule(lr){10-11}\cmidrule(lr){12-13}
                        & Acc & Time & Acc & Time & Acc & Time & Acc & Time & Acc & Time & Acc & Time \\
\midrule
FedAvg \cite{mcmahan2017communication}   & 59.26  & 1.00  &  66.78  & 1.00  &  53.49 & 1.00   &  66.04  & 1.00   &  79.84  & 1.00    &  65.81  & 1.00  \\
\rowcolor{gray!15}
\quad + FLFA             &   \textbf{61.73}  &  1.02  &  \textbf{69.95}   & 1.00     &   \textbf{54.50}  & 1.01     &  \textbf{67.90}   & 1.01     &  \textbf{80.02}   & 1.00     &  \textbf{67.34}   &  1.00  \\
\midrule
FedProx \cite{FedProx}   &   59.67  &   2.01   &   69.28  & 1.98     &  54.00   & 2.01     &   69.03  & 1.96     &   80.75  & 1.99     &  64.88   &   2.08  \\
\rowcolor{gray!15}
\quad + FLFA             &  \textbf{62.46}   &  2.02    &   \textbf{70.41}  & 2.00     &   \textbf{54.22}  & 2.02     &  \textbf{70.29}   & 1.99     &  \textbf{81.56}   & 2.01    &  \textbf{65.86}   & 2.09   \\
\midrule
MOON  \cite{MOON}     &   60.69  & 1.69  &  \textbf{70.98}   & 1.68     &  52.82   & 1.71     &  66.69   & 1.68     &   80.14  & 1.70     &   67.74  &   1.73     \\
\rowcolor{gray!15}
\quad + FLFA             &  \textbf{61.86}   &  1.69  &   70.81  & 1.69     &  \textbf{55.32}   & 1.72     &  \textbf{69.21}   & 1.67     &  \textbf{80.93}   & 1.71     &   \textbf{67.75}  & 1.71    \\
\midrule
FedAvgM \cite{FedAvgM}   &  67.50   & 1.00  &  72.56   & 1.00     &   \textbf{58.49}  & 1.01    &   71.27  & 1.00     &   78.58  & 1.01     &  65.79   & 1.02     \\
\rowcolor{gray!15}
\quad + FLFA             &   \textbf{68.82} &  1.01   &  \textbf{73.28}   & 1.02    &  58.20   & 1.01     &  \textbf{74.39}   & 1.01     &   \textbf{85.03}  & 1.01     &  \textbf{66.27}   & 1.00     \\
\midrule
FedSOL \cite{lee2024fedsol}     &  59.64   &  3.10  &  70.59   & 3.15     &   55.55  & 3.08     &   64.03  & 3.12     &   \textbf{81.38}  & 3.18     &   68.26  & 3.34     \\
\rowcolor{gray!15}
\quad + FLFA             &  \textbf{60.99}   &  3.17    &  \textbf{71.06}   & 3.18     &   \textbf{56.40}  & 3.07     &   \textbf{66.48}  & 3.13     &  81.26   & 3.15     &   \textbf{68.44}  & 3.37    \\
\midrule
FedRS \cite{li2021fedrs}   &   76.31  &  1.08  &   73.07  & 1.09     &  57.73   & 1.08     &  75.55   & 1.11     &   85.59  & 1.07     &  \textbf{69.54}   & 1.10     \\
\rowcolor{gray!15}
\quad + FLFA             &  \textbf{77.41}   &  1.08  &   \textbf{73.85}  & 1.08     &   \textbf{58.52}  & 1.09     &  \textbf{76.77}   & 1.11   &  \textbf{86.07}   & 1.10     &   69.40  & 1.09     \\
\midrule
FedLC \cite{fedlc}      &  75.69   & 1.07   &   73.34  & 1.07     &  58.62   & 1.08     &  75.78   & 1.09     &  85.44   & 1.08     &  68.83   & 1.10     \\
\rowcolor{gray!15}
\quad + FLFA             &  \textbf{77.72}   & 1.09   &   \textbf{73.88}  & 1.07     &   \textbf{59.75}  & 1.07     &  \textbf{77.58}   & 1.11    &  \textbf{86.11}   & 1.11     &  \textbf{69.67}   & 1.10     \\
\midrule
FedACG \cite{fedacg}     &  60.47   &  2.22   &  70.22   & 2.18     &   \textbf{53.90}  & 2.25     &   69.03  & 2.20     &   80.94  & 2.21     &  66.48   & 2.24     \\
\rowcolor{gray!15}
\quad + FLFA             &  \textbf{62.18}   & 2.24    &  \textbf{70.76}   & 2.20    &   53.87  & 2.22     &  \textbf{70.29}   & 2.21     &   \textbf{80.96}  & 2.23     &  \textbf{66.93}   & 2.25     \\
\bottomrule
\end{tabular}
\caption{Test accuracy (\%) and relative training time comparison on various FL algorithms with and without FLFA. Results are averaged over three runs. FLFA achieves performance gains up to 6.5\%p \textbf{with negligible computational overhead}, demonstrating its effectiveness across diverse federated learning algorithms. Best results are in \textbf{bold}.}
\label{tab:accuracy}
\end{table*}

\subsection{Convergence analysis} \label{sec:convergence}
In this section, we establish convergence guarantees for FLFA. We first present our assumptions, then provide convergence results for both local and global models.

\textbf{Assumptions.} For our convergence analysis, we make the following standard assumptions from non-convex optimization literature:
\begin{assumption} \label{asmp:smooth}
The objective function $J_i$ has $M$-Lipschitz continuous gradients:
\begin{equation}
\|\nabla J_i(w) - \nabla J_i(v)\| \leq M\|w - v\|, \quad \forall w, v, \forall i
\end{equation}
\end{assumption}
\begin{assumption} \label{asmp:stoch}
For each client $i$, the stochastic gradients $\nabla \ell(w;x,y)$ are unbiased with bounded variance:
\begin{gather}
    \mathbb{E}_{(x,y) \sim \mathcal{D}_i}[\nabla \ell(w;x,y) | w] = \nabla J_i(w) \\
\mathbb{E}_{(x,y) \sim \mathcal{D}_i}[\|\nabla \ell(w;x,y) - \nabla J_i(w)\|^2 | w] \leq \sigma^2, \quad \forall i    
\end{gather}
\end{assumption}
\begin{assumption} \label{asmp:het}
The divergence between local and global gradients is bounded as
\begin{equation}
\sum_{i} \pi_i \|\nabla J_i(w) - \nabla \mathcal{J}(w)\|^2 \leq \gamma^2
\end{equation}
where $\gamma \geq 0$ are constants and $\pi_i = \frac{|\mathcal{D}_i|}{|\mathcal{D}|}$.
\end{assumption}
\begin{assumption} \label{asmp:fa_error}
For each client $i$ the difference between the gradient computed by FLFA and the true gradient is bounded by:
\begin{equation}
\|\nabla_w^B \ell(w;x,y) - \nabla \ell(w;x,y)\|^2 \leq G^2
\end{equation}
where $G$ depends on the alignment between feedback and weight matrices.
\end{assumption}
Assumptions \ref{asmp:smooth}-\ref{asmp:het} are standard in FL convergence analysis. Assumption~\ref{asmp:fa_error} captures the difference introduced by FLFA. 

\textbf{Local model convergence.} We first establish that individual client models converge despite the FA approximation.
\begin{lem} \label{lem:local_model} For a client $i$ at round $r$ with $\eta < \frac{1}{M}$, the expected decrease in the local objective function after $S$ steps is bounded by:
\begin{equation}
\begin{split}
    \mathbb{E}[J_i(w_i^r) -& J_i(w_i^{r+1})] \geq \eta S(1 - L\eta)\|\nabla J_i(w_i^r)\|^2 \\
    &- \eta SG\|\nabla J_i(w_i^r)\| - \frac{M\eta^2 S}{2}(\sigma^2 + G^2)
\end{split}
\end{equation}
\begin{proof}
    See Section \ref{sec:supp:lem1} in the supplementary materials for the proof.
\end{proof}
\end{lem}
This bound shows that the objective decreases when the gradient norm $\|\nabla J_i(w_i^r)\|^2$ is large enough to overcome the error terms from stochastic noise ($\sigma^2$) and FA approximation ($G^2$). Note that the bound can be negative when the gradient is small---this is expected and desired in non-convex optimization, as it indicates convergence to a neighborhood of a stationary point where gradients vanish. The key insight is that the FA error $G$ contributes to the size of this neighborhood but does not prevent convergence.

\textbf{Global model convergence.} We now establish convergence for the global model aggregated across clients.
\begin{lem}\label{lem:global_decrease} With $\eta < \frac{1}{M}$, the expected decrease in the global objective function after one communication round is bounded by:
\begin{equation}
\begin{split}
    \mathbb{E}[\mathcal{J}(W^r) - \mathcal{J}(W^{r+1})] \geq \frac{\eta S}{4}(1 - M\eta)\|\nabla \mathcal{J}(W^r)\|^2 \\- \eta SG\|\nabla \mathcal{J}(W^r)\| - \frac{M\eta^2 S}{2}(\sigma^2 + G^2 + \gamma^2)
\end{split}
\end{equation}
\begin{proof}
    See Section \ref{sec:supp:lem2} in the supplementary materials for the proof.
\end{proof}
\end{lem}
Similar to the local case, convergence is guaranteed when $\|\nabla J(W^r)\|^2$ is sufficiently large. The bound includes three error sources: stochastic noise ($\sigma^2$), FA approximation ($G^2$), and data heterogeneity ($\gamma^2$). Convergence is to a neighborhood of a stationary point, with $\sigma$, $G$, and $\gamma$ determining the neighborhood size. 

\textbf{Minimizing approximation error $G$.} Using feedback matrices $B$ instead of weight transposes $W^T$ introduces approximation error bounded by $G^2$. Our lemmas show that small $G$ is critical for tight convergence guarantees. Traditional FA uses a random $B$ unrelated to $W$, resulting in large $G$ due to the distance between $B$ and $W$. In contrast, our design choice $B = W^r$ minimizes this error: since $B$ is initialized from the global model, $B$ remains closely aligned with $W$, keeping $G$ small. We validate this design empirically in Section~\ref{sec:abl}, where random $B$ fails to improve over BP. In Section \ref{sec:supp:add_exp} of the supplementary material, we demonstrate that FLFA remains robust even under conditions that could increase $G$, such as increased local steps ($E=15$).

\begin{figure*}[ht!]
\centering
\begin{subfigure}{0.33\textwidth}
    \centering
    \includegraphics[width=\linewidth, trim=0 0 0 0, clip]{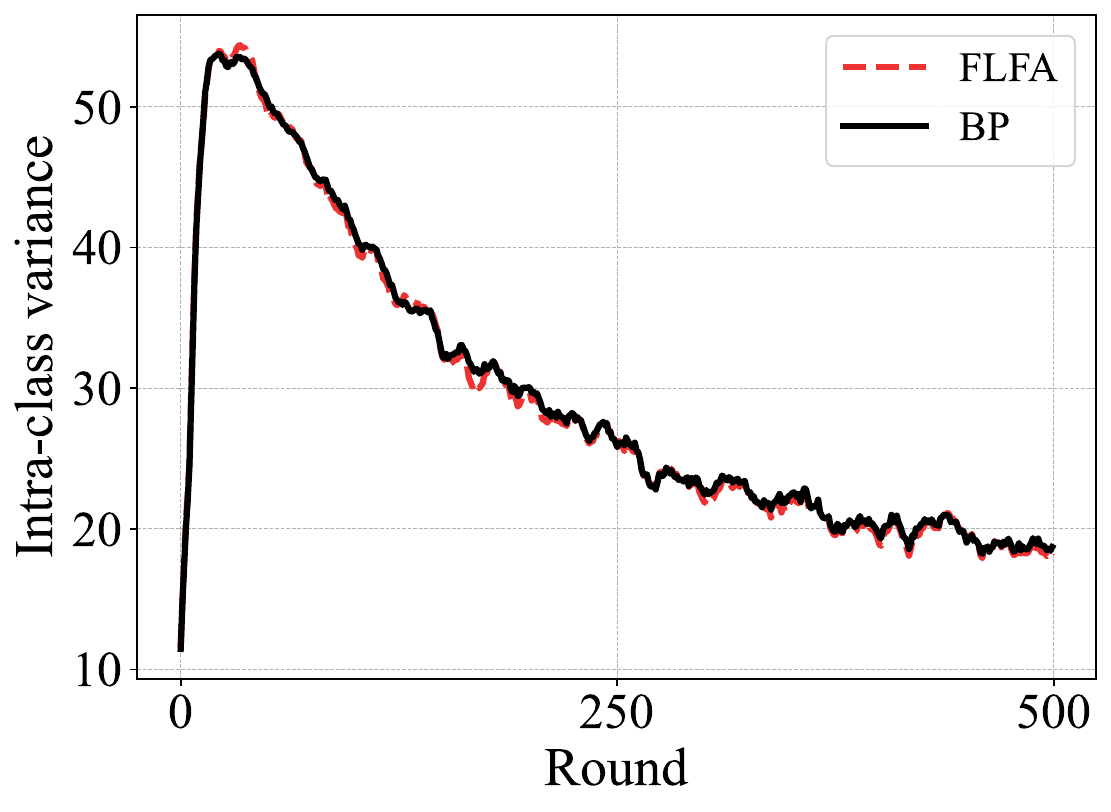}
    \caption{Intra-class variance (lower is better)}
    \label{fig:intra_class_variance}
\end{subfigure}
\hfill
\begin{subfigure}{0.33\textwidth}
    \centering
    \includegraphics[width=\linewidth, trim=0 0 0 0, clip]{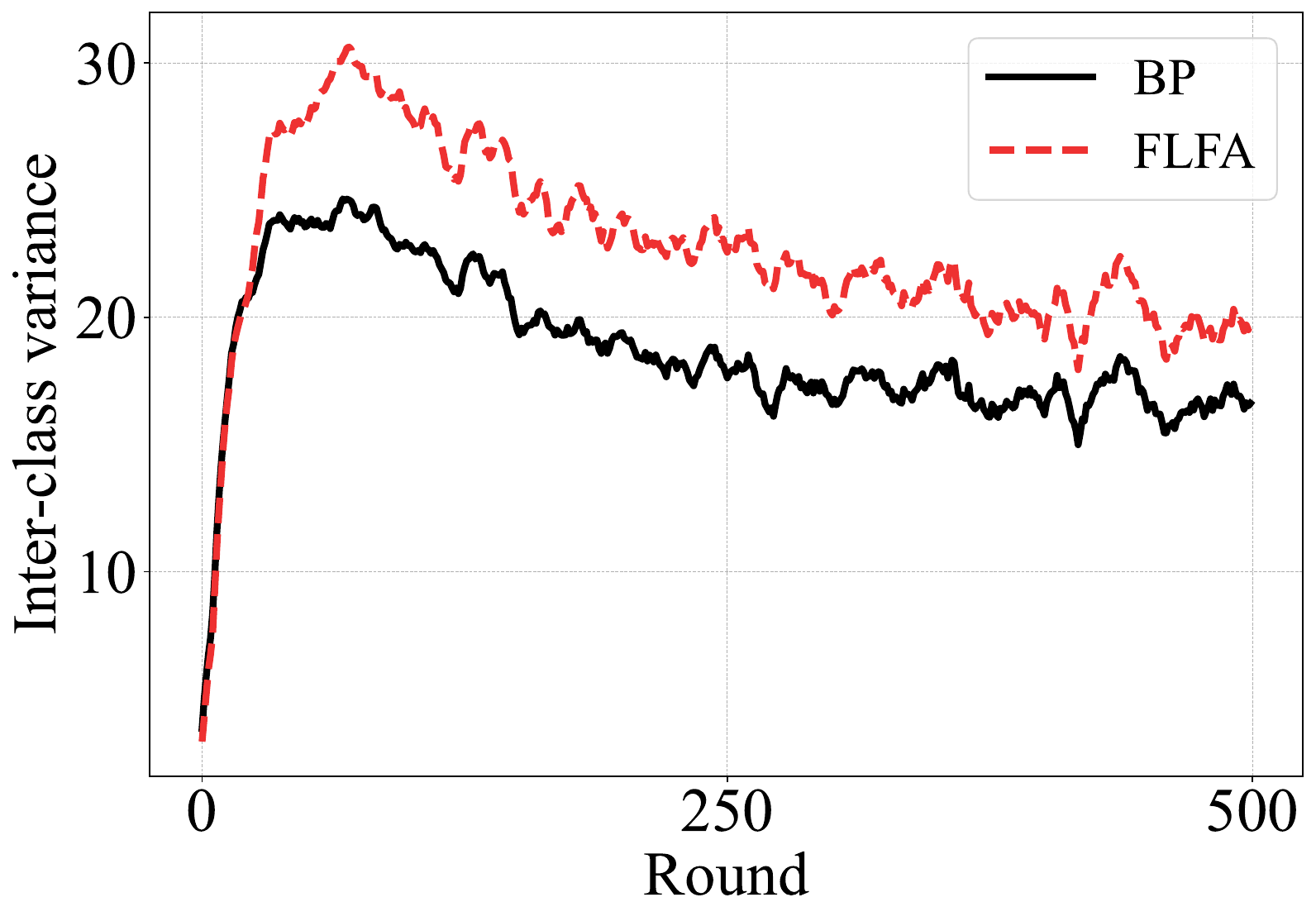}
    \caption{Inter-class variance (higher is better)}
    \label{fig:inter_class_variance}
\end{subfigure}
\hfill
\begin{subfigure}{0.33\textwidth}
    \centering
    \includegraphics[width=\linewidth, trim=0 0 0 0, clip]{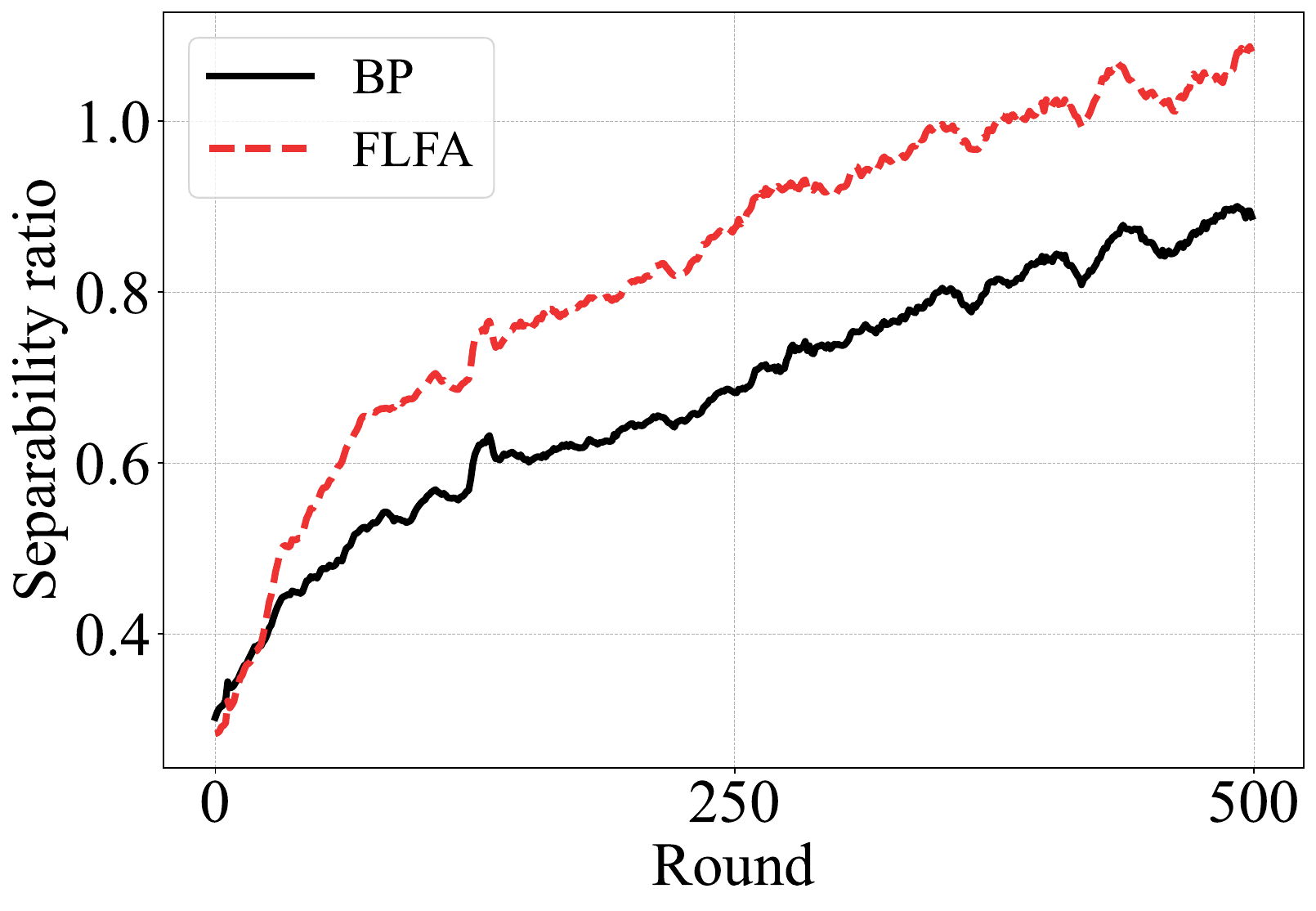}
    \caption{Separability ratio (higher is better)}
    \label{fig:sep_ratio}
\end{subfigure}
\caption{Representation analysis of FLFA vs. BP on CIFAR-10 with FedAvg. FLFA's shared feedback mechanism enables coordinated learning across clients, resulting in consistently higher inter-class separation and 20\% better separability ratio.}
\label{fig:evidence}
\end{figure*}

\section{Experiments}
In this section, we empirically evaluate FLFA through: (1) accuracy comparison against diverse FL methods, (2) model representation and local drift analysis, (3) robustness validation along to diverse settings, and (4) ablation studies validating our design choices.

\textbf{Experimental setup.} We simulate an FL environment with 100 clients, selecting 10\% randomly per round. We use MobileNetV2~\cite{sandler2018mobilenetv2}, a parameter-efficient architecture suitable for FL settings. We evaluate on six datasets: four medical imaging datasets (BloodMNIST, OrganCMNIST, OrganSMNIST, PathMNIST)~\cite{medmnist}, Fashion-MNIST (FMNIST)~\cite{xiao2017fashion}, and CIFAR-10~\cite{krizhevsky2009learning}. We run 100 communication rounds for medical datasets and FMNIST, and 500 rounds for CIFAR-10, with 5 local epochs per round. To simulate realistic non-IID data, we partition data using a Dirichlet distribution with $\beta=0.3$---smaller $\beta$ values indicate higher heterogeneity. Following Section~\ref{sec:discussion}, we apply FLFA to either the layer with highest or lowest gradient similarity. Additional dataset details, data distributions, and hyperparameters are provided in Section \ref{sec:sup_exp_settings} of the supplementary material.

\subsection{Accuracy comparison}
Table~\ref{tab:accuracy} presents the main results, with training time reported as multiples relative to FedAvg. FLFA consistently improves accuracy across 8 FL algorithms and 6 datasets, with results averaged over three random seeds. Notably, FLFA achieves substantial gains: +2.47\% over FedAvg on BloodMNIST, +6.45\% over FedAvgM on FMNIST, and +2.84\% over FedLC on PathMNIST. Even advanced methods like FedRS and FedLC---which already incorporate specialized techniques for non-IID data---benefit from FLFA. 

Importantly, these gains come at negligible computational cost. FLFA introduces small training time overhead compared to base algorithms, as it requires only gradient norm adjustments without auxiliary loss computations or structural changes. This makes FLFA a highly practical, general-purpose module that can be readily integrated into existing FL frameworks.

\subsection{Representation and local drift analysis} 
\label{sec:exp_local_drift}

To understand how FLFA improves performance, we analyze learned feature representations and directly measure local drift using FedAvg on CIFAR-10. For FLFA, we select the layer with the lowest cosine similarity between client gradients to apply FA.

\textbf{Representation quality metrics.} We extract features from the penultimate layer and compute three metrics (formal definitions in Section~\ref{sec:supp:metrics} of the supplementary materials). (1) \textit{Intra-class variance} measures within-class compactness as average distance from samples to their class centroids (lower is better). (2) \textit{Inter-class variance} measures between-class separation as average distance between class centroids (higher is better). (3) \textit{Separability ratio} combines both as inter/intra ratio, capturing overall feature quality (higher is better).

\textbf{Representation quality analysis.} Figure~\ref{fig:intra_class_variance} shows both methods maintain similar intra-class variance, indicating similar within-class compactness. However, Figure~\ref{fig:inter_class_variance} reveals FLFA achieves consistently higher inter-class variance, especially after initial rounds. This means FLFA learns representations where different classes are more separated. Consequently, Figure~\ref{fig:sep_ratio} shows FLFA achieves approximately 20\% higher separability ratios than BP throughout training. This quantitatively confirms our hypothesis: FLFA's shared feedback matrix $B = W^r$ acts as a global anchor, providing clients with structural information about the global objective. While each client learns from its local data in the forward pass, the backward pass guided by $B$ coordinates updates across clients.

\begin{table}[b!]
\small
\centering
\renewcommand{\arraystretch}{1.2}
\begin{tabular}{lccc}
\toprule
\textbf{Dataset} & \textbf{Version} & \textbf{BP} & \textbf{FLFA} \\
\midrule
\multirow{2}{*}{CIFAR-10.1} & V4 & 52.61 & \textbf{54.73} \\
                            & V6 & 52.82 & \textbf{55.16} \\
\bottomrule
\end{tabular}
\caption{Generalization accuracy (\%) on CIFAR-10.1 distribution shift test sets. FLFA outperforms BP, demonstrating robustness to distribution shifts.}
\label{tab:cifar10_1_acc}
\end{table}

\textbf{Generalization to distribution shifts.} We test whether FLFA's improved representations generalize beyond the training distribution using CIFAR-10.1~\cite{recht1806cifar}, which contains new test samples designed to evaluate robustness to distribution shifts. Table~\ref{tab:cifar10_1_acc} shows FLFA outperforms BP by +2.12\% (V4) and +2.34\% (V6), confirming that better feature separation translates to stronger generalization under distribution shifts.

\begin{figure}[t!]
\centering
\includegraphics[width=0.9\columnwidth, trim=0 0 0 0, clip]{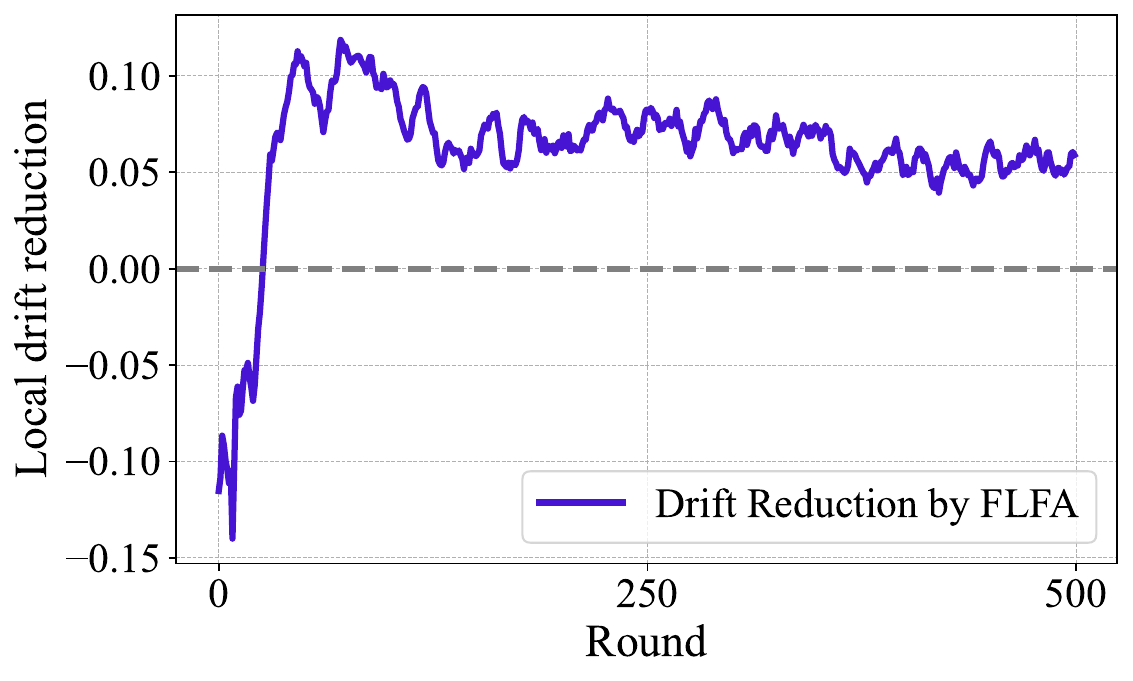}
\caption{Per-round local drift reduction by FLFA over BP, measured as average L2 distance between client updates and their mean. \textbf{Positive values} show FLFA consistently reduces drift, particularly after initial training.}
\label{fig:local_drift_analysis}
\end{figure}

\textbf{Local drift measurement.} Finally, we directly quantify local drift as $\mathcal{H} = \frac{1}{K}\sum_{i=1}^K \|\Delta \mathbf{w}_i - \Delta\Bar{ \mathbf{w}}\|_2$, where $\Delta \mathbf{w}_i$ is client $i$'s flattened weight update and $ \Delta\Bar{ \mathbf{w}}$ is the mean update. Figure~\ref{fig:local_drift_analysis} plots drift reduction per round ($\mathcal{H}_{BP} - \mathcal{H}_{FLFA}$). Positive values indicate FLFA reduces drift more than BP. The result shows FLFA consistently reduces drift, especially after initial rounds when the global model has accumulated meaningful information. This directly validates our theoretical bound (Eq.~\eqref{eq:fa_delta_diff_2}) showing FLFA eliminates the weight divergence term.

\subsection{Robustness validation}
We validate FLFA's effectiveness on more complex datasets and under challenging conditions. While FL research often focuses on simpler datasets, we demonstrate FLFA's robustness on CIFAR-100~\cite{krizhevsky2009learning} (100 classes, 500 rounds) and ImageNet-100~\cite{russakovsky2015imagenet} (100 classes from ImageNet, 100 rounds). Table~\ref{tab:complex_dataset} shows FLFA consistently improves accuracy across all baselines on both datasets. On CIFAR-100, FLFA achieves gains up to +1.10\% (FedAvg), while on ImageNet-100, improvements reach +2.32\% (FedProx, FedACG). These results confirm FLFA generalizes beyond simple benchmarks to challenging scenarios.

\begin{table}[t!]
\small
\centering
\setlength{\tabcolsep}{8pt}
\renewcommand{\arraystretch}{1.1}
\begin{tabular}{l cc cc}
\toprule
\multirow{2}{*}{Method} & \multicolumn{2}{c}{CIFAR-100} & \multicolumn{2}{c}{ImageNet-100} \\
\cmidrule(lr){2-3}\cmidrule(lr){4-5}
& BP & FLFA & BP & FLFA \\
\midrule
FedAvg \cite{mcmahan2017communication} & 42.48 & \textbf{43.58} & 42.84 & \textbf{43.32} \\
FedProx \cite{FedProx} & 42.64 & \textbf{43.27} & 42.30 & \textbf{44.62} \\
MOON \cite{MOON} & 42.62 & \textbf{43.19} & 42.62 & \textbf{44.58} \\
FedAvgM \cite{FedAvgM} & 43.49 & \textbf{43.71} & 43.84 & \textbf{44.24} \\
FedSOL \cite{lee2024fedsol} & 42.88 & \textbf{43.60} & \textbf{43.69} & 43.29 \\
FedRS \cite{li2021fedrs} & 43.98 & \textbf{44.19} & 43.28 & \textbf{43.72} \\
FedLC \cite{fedlc} & 43.85 & \textbf{44.22} & 43.10 & \textbf{44.04} \\
FedACG \cite{fedacg} & 42.64 & \textbf{43.27} & 42.30 & \textbf{44.62} \\
\bottomrule
\end{tabular}
\caption{Test accuracy (\%) on complex datasets. FLFA maintains consistent improvements across diverse baselines, demonstrating scalability to challenging scenarios.}
\label{tab:complex_dataset}
\end{table}

\textbf{Additional robustness tests.} We evaluate FLFA under additional challenging conditions to verify its robustness (full results are in Section \ref{sec:supp:add_exp} of the supplementary materials). First, we test with increased local steps to validate robustness even when the FA approximation error $G$ may accumulate. Second, we examine extreme heterogeneity scenarios with severe non-IID data distributions. Third, we assess scalability using larger network architectures. Finally, we vary client configurations, testing different numbers of participating clients and sampling rates. Across all settings, FLFA maintains consistent performance improvements, empirically validating our theoretical analysis.

\subsection{Ablation study} \label{sec:abl}
We validate FLFA's design choices through ablation studies on CIFAR-10 with FedAvg (\Cref{tab:abl}). FLFA$^*$ replaces our feedback alignment with a fixed random feedback matrix (standard FA). FLFA$^\dagger$ removes the adaptive scaling mechanism from \Cref{sec:theoretical_foundation}. For all experiments, we select the layer with the lowest cosine similarity between client gradients to apply FA.

\begin{table}[b!]
\small
\centering
\begin{tabular}{ccccc}
\toprule
\textbf{BP} & \textbf{FLFA} & \textbf{FLFA$^*$} & \textbf{FLFA$^\dagger$} & \textbf{DFA} \\
\midrule
65.81 & \textbf{68.05} & 64.02 & 65.21 & - \\
\bottomrule
\end{tabular}
\caption{Ablation study on FLFA components. FLFA$^*$: random feedback matrix; FLFA$^\dagger$: no adaptive scaling; DFA: direct feedback alignment.}
\label{tab:abl}
\end{table}

\textbf{Global weights vs. random feedback (FLFA$^*$).} Replacing $B = W^r$ with a fixed random matrix degrades performance below the BP baseline. This confirms that using global weights minimizes approximation error $G$ through alignment with local weights, while random feedback fails to coordinate client updates effectively.

\textbf{Adaptive scaling (FLFA$^\dagger$).} Removing adaptive scaling reduces FLFA's gain from +2.24\% to -0.60\%, which is worse than BP. This validates our theoretical analysis (Eq.~\eqref{eq:fa_delta_diff_2}): adaptive scaling maintains $\|B\| = \|w\|$, ensuring the term $|1-\alpha|\|B\|$ remains negligible. Without it, the feedback matrix becomes misaligned with local weight.

\textbf{DFA.} We also tested DFA \cite{nokland2016direct}, where each layer receives feedback directly from the output via random matrices. Similar to FLFA, we select the layer with the lowest cosine similarity to apply DFA. However, DFA fails to converge on this task. This is consistent with prior work showing that DFA is primarily suitable for simple or linear models and is known to perform poorly on CNNs.

\textbf{Layer selection studies.} FLFA's only hyperparameter is which layer to apply feedback alignment. We find that applying FA to a single layer is sufficient for effective drift reduction. In Section~\ref{sec:supp:add_abl} of the supplementary materials, we provide comprehensive analyses on: (1) the performance of applying FA to each individual layer, (2) the impact of applying FA to multiple layers simultaneously, and (3) guidelines for layer selection based on gradient similarity between clients---specifically, when to select the most aligned layer (lowest disagreement) versus the most divergent layer (highest disagreement) depending on the overall drift level.

\section{Conclusion}
In this paper, we introduced \textbf{FLFA (Federated Learning with Feedback Alignment)}, which modifies gradient computation using the global model's weights as a shared feedback matrix during backpropagation, instead of constraining model training through regularization. This creates a unified backward pathway that aligns local updates across heterogeneous clients without auxiliary objectives.

Our work provides both theoretical foundations and empirical validation for this approach. Theoretically, we established tighter drift bounds and convergence guarantees demonstrating why FLFA effectively mitigates client divergence. Empirically, we demonstrated consistent improvements across diverse FL algorithms, datasets, and settings.

Most importantly, FLFA operates with zero communication overhead and minimal computational cost. Its seamless integration ensures that it is easily deployable. FLFA advances the FL state-of-the-art by rethinking fundamental training mechanisms instead of adding architectural complexity. This work demonstrates that addressing core algorithmic principles can yield more practical solutions than increasing system complexity.


{
    \small
    \bibliographystyle{ieeenat_fullname}
    \bibliography{main}

@article{lillicrap2016random,
  title={Random synaptic feedback weights support error backpropagation for deep learning},
  author={Lillicrap, Timothy P and Cownden, Daniel and Tweed, Douglas B and Akerman, Colin J},
  journal={Nature communications},
  volume={7},
  number={1},
  pages={13276},
  year={2016},
  publisher={Nature Publishing Group UK London}
}

@inproceedings{mcmahan2017communication,
  title={Communication-efficient learning of deep networks from decentralized data},
  author={McMahan, Brendan and Moore, Eider and Ramage, Daniel and Hampson, Seth and y Arcas, Blaise Aguera},
  booktitle={Artificial intelligence and statistics},
  pages={1273--1282},
  year={2017},
  organization={PMLR}
}

@inproceedings{MOON,
  title={Model-contrastive federated learning},
  author={Li, Qinbin and He, Bingsheng and Song, Dawn},
  booktitle={Proceedings of the IEEE/CVF conference on computer vision and pattern recognition},
  pages={10713--10722},
  year={2021}
}

@article{FedProx,
  title={Federated optimization in heterogeneous networks},
  author={Li, Tian and Sahu, Anit Kumar and Zaheer, Manzil and Sanjabi, Maziar and Talwalkar, Ameet and Smith, Virginia},
  journal={Proceedings of Machine learning and systems},
  volume={2},
  pages={429--450},
  year={2020}
}

@article{FedAvgM,
  title={Measuring the effects of non-identical data distribution for federated visual classification},
  author={Hsu, Tzu-Ming Harry and Qi, Hang and Brown, Matthew},
  journal={arXiv preprint arXiv:1909.06335},
  year={2019}
}

@inproceedings{he2016deep,
  title={Deep residual learning for image recognition},
  author={He, Kaiming and Zhang, Xiangyu and Ren, Shaoqing and Sun, Jian},
  booktitle={Proceedings of the IEEE conference on computer vision and pattern recognition},
  pages={770--778},
  year={2016}
}

@article{xiao2017fashion,
  title={Fashion-mnist: a novel image dataset for benchmarking machine learning algorithms},
  author={Xiao, Han and Rasul, Kashif and Vollgraf, Roland},
  journal={arXiv preprint arXiv:1708.07747},
  year={2017}
}

@article{krizhevsky2009learning,
  title={Learning multiple layers of features from tiny images},
  author={Krizhevsky, Alex and Hinton, Geoffrey and others},
  year={2009},
  publisher={Toronto, ON, Canada}
}

@article{jung2023lafd,
  title={LAFD: Local-differentially private and asynchronous federated learning with direct feedback alignment},
  author={Jung, Kijung and Baek, Incheol and Kim, Soohyung and Chung, Yon Dohn},
  journal={IEEE Access},
  volume={11},
  pages={86754--86769},
  year={2023},
  publisher={IEEE}
}

@inproceedings{colombo2023tifed,
  title={TIFeD: a Tiny Integer-based Federated learning algorithm with Direct feedback alignment},
  author={Colombo, Luca and Falcetta, Alessandro and Roveri, Manuel},
  booktitle={Proceedings of the Third International Conference on AI-ML Systems},
  pages={1--8},
  year={2023}
}

@article{nokland2016direct,
  title={Direct feedback alignment provides learning in deep neural networks},
  author={N{\o}kland, Arild},
  journal={Advances in neural information processing systems},
  volume={29},
  year={2016}
}

@article{russakovsky2015imagenet,
  title={Imagenet large scale visual recognition challenge},
  author={Russakovsky, Olga and Deng, Jia and Su, Hao and Krause, Jonathan and Satheesh, Sanjeev and Ma, Sean and Huang, Zhiheng and Karpathy, Andrej and Khosla, Aditya and Bernstein, Michael and others},
  journal={International journal of computer vision},
  volume={115},
  number={3},
  pages={211--252},
  year={2015},
  publisher={Springer}
}

@inproceedings{lee2024fedsol,
  title={Fedsol: Stabilized orthogonal learning with proximal restrictions in federated learning},
  author={Lee, Gihun and Jeong, Minchan and Kim, Sangmook and Oh, Jaehoon and Yun, Se-Young},
  booktitle={Proceedings of the IEEE/CVF Conference on Computer Vision and Pattern Recognition},
  pages={12512--12522},
  year={2024}
}

@inproceedings{li2021fedrs,
  title={Fedrs: Federated learning with restricted softmax for label distribution non-iid data},
  author={Li, Xin-Chun and Zhan, De-Chuan},
  booktitle={Proceedings of the 27th ACM SIGKDD conference on knowledge discovery \& data mining},
  pages={995--1005},
  year={2021}
}

@inproceedings{sandler2018mobilenetv2,
  title={Mobilenetv2: Inverted residuals and linear bottlenecks},
  author={Sandler, Mark and Howard, Andrew and Zhu, Menglong and Zhmoginov, Andrey and Chen, Liang-Chieh},
  booktitle={Proceedings of the IEEE conference on computer vision and pattern recognition},
  pages={4510--4520},
  year={2018}
}

@article{recht1806cifar,
  title={Do cifar-10 classifiers generalize to cifar-10? 2018},
  author={Recht, Benjamin and Roelofs, Rebecca and Schmidt, Ludwig and Shankar, Vaishaal},
  journal={URL https://arxiv. org/abs},
  year={1806}
}

@inproceedings{fedlc,
  title={Federated learning with label distribution skew via logits calibration},
  author={Zhang, Jie and Li, Zhiqi and Li, Bo and Xu, Jianghe and Wu, Shuang and Ding, Shouhong and Wu, Chao},
  booktitle={International Conference on Machine Learning},
  pages={26311--26329},
  year={2022},
  organization={PMLR}
}

@inproceedings{fedacg,
  title={Communication-efficient federated learning with accelerated client gradient},
  author={Kim, Geeho and Kim, Jinkyu and Han, Bohyung},
  booktitle={Proceedings of the IEEE/CVF Conference on Computer Vision and Pattern Recognition},
  pages={12385--12394},
  year={2024}
}

@inproceedings{medmnist,
  title={Medmnist classification decathlon: A lightweight automl benchmark for medical image analysis},
  author={Yang, Jiancheng and Shi, Rui and Ni, Bingbing},
  booktitle={2021 IEEE 18th International Symposium on Biomedical Imaging (ISBI)},
  pages={191--195},
  year={2021},
  organization={IEEE}
}

@article{zhao2018federated,
  title={Federated learning with non-iid data},
  author={Zhao, Yue and Li, Meng and Lai, Liangzhen and Suda, Naveen and Civin, Damon and Chandra, Vikas},
  journal={arXiv preprint arXiv:1806.00582},
  year={2018}
}

@article{li2019convergence,
  title={On the convergence of fedavg on non-iid data},
  author={Li, Xiang and Huang, Kaixuan and Yang, Wenhao and Wang, Shusen and Zhang, Zhihua},
  journal={arXiv preprint arXiv:1907.02189},
  year={2019}
}

@inproceedings{karimireddy2020scaffold,
  title={Scaffold: Stochastic controlled averaging for federated learning},
  author={Karimireddy, Sai Praneeth and Kale, Satyen and Mohri, Mehryar and Reddi, Sashank and Stich, Sebastian and Suresh, Ananda Theertha},
  booktitle={International conference on machine learning},
  pages={5132--5143},
  year={2020},
  organization={PMLR}
}

@article{wang2020tackling,
  title={Tackling the objective inconsistency problem in heterogeneous federated optimization},
  author={Wang, Jianyu and Liu, Qinghua and Liang, Hao and Joshi, Gauri and Poor, H Vincent},
  journal={Advances in neural information processing systems},
  volume={33},
  pages={7611--7623},
  year={2020}
}

@article{acar2021federated,
  title={Federated learning based on dynamic regularization},
  author={Acar, Durmus Alp Emre and Zhao, Yue and Navarro, Ramon Matas and Mattina, Matthew and Whatmough, Paul N and Saligrama, Venkatesh},
  journal={arXiv preprint arXiv:2111.04263},
  year={2021}
}

@article{wang2020federated,
  title={Federated learning with matched averaging},
  author={Wang, Hongyi and Yurochkin, Mikhail and Sun, Yuekai and Papailiopoulos, Dimitris and Khazaeni, Yasaman},
  journal={arXiv preprint arXiv:2002.06440},
  year={2020}
}

@article{sattler2019robust,
  title={Robust and communication-efficient federated learning from non-iid data},
  author={Sattler, Felix and Wiedemann, Simon and M{\"u}ller, Klaus-Robert and Samek, Wojciech},
  journal={IEEE transactions on neural networks and learning systems},
  volume={31},
  number={9},
  pages={3400--3413},
  year={2019},
  publisher={IEEE}
}

@inproceedings{mohri2019agnostic,
  title={Agnostic federated learning},
  author={Mohri, Mehryar and Sivek, Gary and Suresh, Ananda Theertha},
  booktitle={International conference on machine learning},
  pages={4615--4625},
  year={2019},
  organization={PMLR}
}
}

\clearpage
\setcounter{page}{1}
\maketitlesupplementary

\section{Detailed derivation of Section 4.1} \label{sec:supp:derivation}
In this section, we provide a detailed derivation of the upper bound for $\Vert \Delta^r_{i, l} - \Delta^r_{j, l} \Vert_F $. The difference between the updates is:
\begin{equation}
    \Delta^r_{i, l} - \Delta^r_{j, l} = -\eta \sum_{k=0}^{s-1} [\delta_{i,l}^{(r,k)} (h_{i,l}^{(r,k)})^T - \delta_{j,l}^{(r,k)} (h_{j,l}^{(r,k)})^T  ]
\end{equation}
Taking the Frobenius norm and applying the triangle inequality:
\begin{equation}
\begin{split}
    & \Vert \Delta^r_{i, l} - \Delta^r_{j, l} \Vert_F \\
    &= \eta \Vert \sum_{k=0}^{s-1} [\delta_{i,l}^{(r,k)} (h_{i,l}^{(r,k)})^T - \delta_{j,l}^{(r,k)} (h_{j,l}^{(r,k)})^T  ] \Vert_F \\
    &\leq  \eta \sum_{k=0}^{s-1} \Vert\delta_{i,l}^{(r,k)} (h_{i,l}^{(r,k)})^T - \delta_{j,l}^{(r,k)} (h_{j,l}^{(r,k)})^T \Vert_F 
\end{split}
\end{equation}
Decompose the term inside the sum, it can be bounded as follows:
\begin{equation} \label{eq:supp:delta_diff}
\begin{split}
    &\Vert\delta_{i,l}^{(r,k)} (h_{i,l}^{(r,k)})^T - \delta_{j,l}^{(r,k)} (h_{j,l}^{(r,k)})^T \Vert_F\\
    &= \Vert (\delta_{i,l}^{(r,k)} - \delta_{j,l}^{(r,k)}) (h_{i,l}^{(r,k)})^T  + \delta_{j,l}^{(r,k)} (h_{i,l}^{(r,k)} - h_{j,l}^{(r,k)})^T  \Vert_F\\
    &\leq  \Vert (\delta_{i,l}^{(r,k)} - \delta_{j,l}^{(r,k)}) (h_{i,l}^{(r,k)})^T \Vert_F + \Vert \delta_{j,l}^{(r,k)} (h_{i,l}^{(r,k)} - h_{j,l}^{(r,k)})^T  \Vert_F\\
    &= \Vert \delta_{i,l}^{(r,k)} - \delta_{j,l}^{(r,k)}\Vert \cdot \Vert h_{i,l}^{(r,k)} \Vert + \Vert \delta_{j,l}^{(r,k)} \Vert \cdot \Vert h_{i,l}^{(r,k)} - h_{j,l}^{(r,k)}  \Vert
\end{split}
\end{equation}
Substituting back:
\begin{equation} \label{eq:supp:detla_diff_2}
    \begin{split}
        & \Vert \Delta^r_{i, l} - \Delta^r_{j, l} \Vert_F \\
        &\leq  \eta \sum_{k=0}^{s-1} \big[ \Vert \delta_{i,l}^{(r,k)} - \delta_{j,l}^{(r,k)} \Vert \cdot \Vert h_{i,l}^{(r,k)} \Vert \\
        &\quad + \Vert \delta_{j,l}^{(r,k)} \Vert \cdot \Vert h_{i,l}^{(r,k)} - h_{j,l}^{(r,k)}  \Vert \big]
    \end{split}
\end{equation}

Now, we derive bounds for $ \Vert \delta_{i,l}^{(r,k)} - \delta_{j,l}^{(r,k)} \Vert$ and $\Vert \delta_{j,l}^{(r,k)} \Vert$. For the first one:
\begin{equation}
    \begin{split}
        & \Vert \delta_{i,l}^{(r,k)} - \delta_{j,l}^{(r,k)} \Vert \\
        & = \big\Vert ((w_{i,l+1}^{(r,k)})^T \delta_{i,l+1}^{(r,k)})  \odot f'(z_{i,l}^{(r,k)}) \\
        &\quad\quad - ((w_{j,l+1}^{(r,k)})^T \delta_{j,l+1}^{(r,k)})  \odot f'(z_{j,l}^{(r,k)}) \big\Vert \\
        &= \big\Vert \big[ (w_{i,l+1}^{(r,k)})^T \delta_{i,l+1}^{(r,k)} - (w_{j,l+1}^{(r,k)})^T \delta_{j,l+1}^{(r,k)}   \big] \odot f'(z_{i,l}^{(r,k)}) \\
        &\quad + \big( (w_{j,l+1}^{(r,k)})^T \delta_{j,l+1}^{(r,k)} \big) \odot \big( f'(z_{i,l}^{(r,k)}) - f'(z_{j,l}^{(r,k)}) \big) \big\Vert \\
        &\leq \big\Vert \big[ (w_{i,l+1}^{(r,k)})^T \delta_{i,l+1}^{(r,k)} - (w_{j,l+1}^{(r,k)})^T \delta_{j,l+1}^{(r,k)}   \big] \odot f'(z_{i,l}^{(r,k)}) \big\Vert \\
        &\quad + \big\Vert \big( (w_{j,l+1}^{(r,k)})^T \delta_{j,l+1}^{(r,k)} \big) \odot \big( f'(z_{i,l}^{(r,k)}) - f'(z_{j,l}^{(r,k)}) \big) \big\Vert
    \end{split}
\end{equation}
Using the property of the Hadamard product, \(\|a \odot b\| \leq \|b\|_\infty \|a\|\), the first term of bound is bounded as:
\begin{equation}
\begin{split}
&\left\| \left[ (w_{i,l+1}^{(r,k)})^T \delta_{i,l+1}^{(r,k)} - (w_{j,l+1}^{(r,k)})^T \delta_{j,l+1}^{(r,k)} \right] \odot f'(z_{i,l}^{(r,k)}) \right\| \\
&\leq \|f'(z_{i,l}^{(r,k)})\|_\infty \left\| (w_{i,l+1}^{(r,k)})^T \delta_{i,l+1}^{(r,k)} - (w_{j,l+1}^{(r,k)})^T \delta_{j,l+1}^{(r,k)} \right\|
\end{split}
\end{equation}
Decompose the inner term:
\begin{equation}
\begin{split}
& \big\Vert(w_{i,l+1}^{(r,k)})^T \delta_{i,l+1}^{(r,k)} - (w_{j,l+1}^{(r,k)})^T \delta_{j,l+1}^{(r,k)}\big\Vert \\
&= \big\Vert(w_{i,l+1}^{(r,k)})^T (\delta_{i,l+1}^{(r,k)} - \delta_{j,l+1}^{(r,k)}) \\
&\quad + \big\Vert \big[ (w_{i,l+1}^{(r,k)})^T - (w_{j,l+1}^{(r,k)})^T \big] \delta_{j,l+1}^{(r,k)} \big\Vert \\
& \leq  \big\Vert w_{i,l+1}^{(r,k)}\big\Vert_2 \big\Vert\delta_{i,l+1}^{(r,k)} - \delta_{j,l+1}^{(r,k)}\big\Vert \\
&\quad + \big\Vert w_{i,l+1}^{(r,k)} - w_{j,l+1}^{(r,k)}\big\Vert_2 \big\Vert\delta_{j,l+1}^{(r,k)}\big\Vert
\end{split}
\end{equation}
Thus:
\begin{equation}
\begin{split}
&\big\Vert \big[ (w_{i,l+1}^{(r,k)})^T \delta_{i,l+1}^{(r,k)} - (w_{j,l+1}^{(r,k)})^T \delta_{j,l+1}^{(r,k)} \big] \odot f'(z_{i,l}^{(r,k)}) \big\Vert \\
&\leq \Vert f'(z_{i,l}^{(r,k)})\Vert_\infty \big[ \big\Vert w_{i,l+1}^{(r,k)}\big\Vert \cdot \big\Vert\delta_{i,l+1}^{(r,k)} - \delta_{j,l+1}^{(r,k)}\big\Vert \\
&\quad + \big\Vert w_{i,l+1}^{(r,k)} - w_{j,l+1}^{(r,k)}\big\Vert \cdot \big\Vert\delta_{j,l+1}^{(r,k)}\big\Vert \big]
\end{split}
\end{equation}
For the second term of bound:
\begin{equation}
\begin{split}
&\left\| \left( (w_{j,l+1}^{(r,k)})^T \delta_{j,l+1}^{(r,k)} \right) \odot \left( f'(z_{i,l}^{(r,k)}) - f'(z_{j,l}^{(r,k)}) \right) \right\| \\
&\leq \left\| f'(z_{i,l}^{(r,k)}) - f'(z_{j,l}^{(r,k)}) \right\|_\infty \left\| (w_{j,l+1}^{(r,k)})^T \delta_{j,l+1}^{(r,k)} \right\| \\
&\leq \left\| f'(z_{i,l}^{(r,k)}) - f'(z_{j,l}^{(r,k)}) \right\|_\infty \|w_{j,l+1}^{(r,k)}\| \cdot \|\delta_{j,l+1}^{(r,k)}\|
\end{split}
\end{equation}
Combine both terms:
\begin{equation}
\begin{split}
&\Vert\delta_{i,l}^{(r,k)} - \delta_{j,l}^{(r,k)}\Vert \\
&\leq \Vert f'(z_{i,l}^{(r,k)})\Vert_\infty \big[ \Vert w_{i,l+1}^{(r,k)}\Vert \cdot \Vert\delta_{i,l+1}^{(r,k)} - \delta_{j,l+1}^{(r,k)}\Vert \\
&\quad + \Vert w_{i,l+1}^{(r,k)} - w_{j,l+1}^{(r,k)}\Vert\cdot \Vert\delta_{j,l+1}^{(r,k)}\Vert \big] \\
&\quad + \big\Vert f'(z_{i,l}^{(r,k)}) - f'(z_{j,l}^{(r,k)})\big\Vert_\infty \Vert w_{j,l+1}^{(r,k)}\Vert\cdot \Vert\delta_{j,l+1}^{(r,k)}\Vert
\end{split}
\end{equation}
A bound for $\Vert \delta_{j,l}^{(r,k)} \Vert$ of \eqref{eq:supp:detla_diff_2} can be derived by:
\begin{equation}
    \begin{split}
        &\Vert \delta_{j,l}^{(r,k)} \Vert = \Vert ((w_{j,l+1}^{(r,k)})^T \delta_{j,l+1}^{(r,k)})  \odot f'(z_{j,l}^{(r,k)}) \Vert \\
        &\leq \Vert f'(z_{j,l}^{(r,k)}) \Vert_\infty \Vert w_{j,l+1}^{(r,k)} \Vert\cdot \Vert \delta_{j,l+1}^{(r,k)} \Vert
    \end{split}
\end{equation}
Substituting back, the final bound for $\Vert \Delta^r_{i, l} - \Delta^r_{j, l} \Vert_F$ is:
\begin{equation}
    \begin{split}
&\Vert \Delta^r_{i, l} - \Delta^r_{j, l} \Vert_F \\
&\leq \eta \sum_{k=0}^{s-1} \big[ \Vert f'(z_{i,l}^{(r,k)})\Vert_\infty \big( \Vert w_{i,l+1}^{(r,k)}\Vert\cdot \Vert\delta_{i,l+1}^{(r,k)} - \delta_{j,l+1}^{(r,k)}\Vert \\
&\quad + \Vert w_{i,l+1}^{(r,k)} - w_{j,l+1}^{(r,k)}\Vert\cdot \Vert\delta_{j,l+1}^{(r,k)}\Vert \big) \Vert h_{i,l}^{(r,k)} \Vert \\
&\quad + \big\Vert f'(z_{i,l}^{(r,k)}) - f'(z_{j,l}^{(r,k)})\big\Vert_\infty \Vert h_{i,l}^{(r,k)} - h_{j,l}^{(r,k)}  \Vert \\
&\quad \quad \cdot \Vert w_{j,l+1}^{(r,k)}\Vert\cdot \Vert\delta_{j,l+1}^{(r,k)}\Vert  \big]
    \end{split}
\end{equation}
For simplicity and to highlight local drift, we assume that
\begin{equation}
   \Vert h_{i,l+1}^{(r,k)} \Vert \leq \Tilde{x}, \quad  \Vert \delta_{i,l+1}^{(r,k)} \Vert \leq \Tilde{\delta}
\end{equation}
for across all clients. Then,
\begin{equation}
    \begin{split}
&\Vert \Delta^r_{i, l} - \Delta^r_{j, l} \Vert_F \\
&\leq \eta \sum_{k=0}^{s-1} \big[ \Tilde{x} \Vert w_{i,l+1}^{(r,k)}\Vert \cdot \Vert\delta_{i,l+1}^{(r,k)} - \delta_{j,l+1}^{(r,k)}\Vert \\
& + \Tilde{x}\Tilde{\delta} \Vert w_{i,l+1}^{(r,k)} - w_{j,l+1}^{(r,k)}\Vert +  \Tilde{\delta}\Vert w_{j,l+1}^{(r,k)}\Vert\cdot \Vert h_{i,l}^{(r,k)} - h_{j,l}^{(r,k)}  \Vert]
    \end{split}
\end{equation}

\section{Algorithm of FLFA} \label{sec:supp:alg}
\begin{algorithm}
\caption{FLFA} \label{alg:FLFA}
    \begin{algorithmic}[1]
    \Require $E, R, \eta, \mathcal{F}$
    \Ensure $\mathbf{W}^R$

    \State Initialize global model $\mathbf{W}^0$
    \For{each round $r=0, \ldots, R-1$}
        \For{each client $ i \in \mathcal{U}^r$}
            \State Initialize local model: $\mathbf{w}_i^r \gets \mathbf{W}^r$
            \For {$l \in \mathcal{F}$}
                \State Initialize feedback matrix: $B_{i,l}^r \gets W_l^r$
            \EndFor
            \For{each epoch $e=0,\ldots,E-1$}
                \For {each batch $(x,y)$}
                    \State \parbox[t]{\linewidth}{%
                      Compute gradients\\
                       using BP for $l \notin \mathcal{F}$ and FA for $l \in \mathcal{F}$\\%
                    }
                    \State Update local model: \[
                    \mathbf{w}_i^r \gets \mathbf{w}_i^r - \eta \cdot \nabla_{\mathbf{w}_i^r}^{\mathbf{B}_i^r} \mathcal{L}(\mathbf{w}_i^r; x, y)
                    \]
                    \For {$l \in \mathcal{F}$}
                        \State Adjust feedback matrix: \[
                        B_{i,l}^r \gets \frac{\Vert w_{i,l}^r \Vert}{\Vert W_{l}^r \Vert}  W_{l}^r
                        \]
                    \EndFor
                \EndFor
            \EndFor
        \EndFor
        \State Aggregate local models: \[
        \mathbf{W}^{r+1} \gets  \frac{1}{\sum_{i \in \mathcal{U}^r} |\mathcal{D}_i| } \sum_{i \in \mathcal{U}^r} |\mathcal{D}_i|\,\mathbf{w}_i^r
        \]
    \EndFor

    \State \Return $\mathbf{W}^R$
    \end{algorithmic}
\end{algorithm}

\section{Deferred Proofs for Lemma 1 and 2}


\subsection{Proof for Lemma 1} \label{sec:supp:lem1}
\begin{proof}
Let $w_i^{r,k}$ denote the local model of client $i$ at round $r$ and step $k$. The update rule is:
\begin{equation}
w_i^{r,k+1} = w_i^{r,k} - \eta \nabla_{w_i^{r,k}}^{B_i^r} \ell(w_i^{r,k};x,y)
\end{equation}

Let $g_i^{r,k} = \nabla \ell(w_i^{r,k};x,y)$ be the true stochastic gradient and $\hat{g}_i^{r,k} = \nabla_{w_i^{r,k}}^{B_i^r} \ell(w_i^{r,k};x,y)$ be the FA gradient.

By Assumption \ref{asmp:smooth} ($M$-Lipschitz):
\begin{equation}
\begin{split}
    J_i(w_i^{r,k+1}) \leq J_i(w_i^{r,k}) + \langle \nabla J_i(w_i^{r,k}), w_i^{r,k+1} - w_i^{r,k} \rangle \\ + \frac{M}{2}\|w_i^{r,k+1} - w_i^{r,k}\|^2
\end{split}
\end{equation}

Substituting the update rule:
\begin{equation}
\begin{split}
    J_i(w_i^{r,k+1}) \leq J_i(w_i^{r,k}) - \eta \langle \nabla J_i(w_i^{r,k}), \hat{g}_i^{r,k} \rangle \\+ \frac{L\eta^2}{2}\|\hat{g}_i^{r,k}\|^2
\end{split}
\end{equation}

Using the inner product decomposition:
\begin{equation}
\begin{split}
    &\langle \nabla J_i(w_i^{r,k}), \hat{g}_i^{r,k} \rangle \\
&= \langle \nabla J_i(w_i^{r,k}), g_i^{r,k} \rangle + \langle \nabla J_i(w_i^{r,k}), \hat{g}_i^{r,k} - g_i^{r,k} \rangle
\end{split}
\end{equation}

By Cauchy-Schwarz and Assumption \ref{asmp:fa_error}:
\begin{equation}
\begin{split}
    \langle \nabla J_i(w_i^{r,k}), \hat{g}_i^{r,k} - g_i^{r,k} \rangle &\geq -\|\nabla J_i(w_i^{r,k})\| \cdot \|\hat{g}_i^{r,k} - g_i^{r,k}\| \\
&\geq -\|\nabla J_i(w_i^{r,k})\| \cdot G
\end{split}
\end{equation}

For the squared norm term:
\begin{equation}
\begin{split}
    \|\hat{g}_i^{r,k}\|^2 &\leq 2\|g_i^{r,k}\|^2 + 2\|\hat{g}_i^{r,k} - g_i^{r,k}\|^2 \\
    &\leq 2\|g_i^{r,k}\|^2 + 2G^2
\end{split}
\end{equation}

Taking expectation with respect to the stochastic gradient and using Assumption \ref{asmp:stoch}:
\begin{equation}
\mathbb{E}_{(x,y) \sim D_i}[\langle \nabla J_i(w_i^{r,k}), g_i^{r,k} \rangle | w_i^{r,k}] = \|\nabla J_i(w_i^{r,k})\|^2
\end{equation}
\begin{equation}
\mathbb{E}_{(x,y) \sim D_i}[\|g_i^{r,k}\|^2 | w_i^{r,k}] = \|\nabla J_i(w_i^{r,k})\|^2 + \sigma^2
\end{equation}

Substituting and summing over $S$ steps:
\begin{equation}
\begin{split}
    \mathbb{E}[J_i(w_i^{r+1})]&\leq \mathbb{E}[J_i(w_i^r)] - \eta S(1 - M\eta)\|\nabla J_i(w_i^r)\|^2 \\
&+ \eta SG\|\nabla J_i(w_i^r)\| + \frac{M\eta^2 S}{2}(\sigma^2 + G^2)
\end{split}
\end{equation}

Rearranging:
\begin{equation}
\begin{split}
    \mathbb{E}[J_i(w_i^r) - J_i(w_i^{r+1})] \geq \eta S(1 - M\eta)\|\nabla J_i(w_i^r)\|^2  \\
 - \eta SG\|\nabla J_i(w_i^r)\| - \frac{M\eta^2 S}{2}(\sigma^2 + G^2)
\end{split}
\end{equation}

For $\eta < \frac{1}{M}$, the coefficient of $\|\nabla J_i(w_i^r)\|^2$ is positive, indicating that the local model converges to a neighborhood of a stationary point, where the neighborhood size depends on $G$, $\sigma$, and $\eta$.
\end{proof}

\subsection{Proof for Lemma 2} \label{sec:supp:lem2}
\begin{proof}
The global model update is:
\begin{equation}
W^{r+1} = \sum_{i} \pi_i w_i^{r+1}
\end{equation}

By Assumption \ref{asmp:smooth} (M-Lipschitz):

\begin{equation}
\begin{split}
\mathcal{J}(W^{r+1})&  \leq \mathcal{J}(W^r) + \langle \nabla \mathcal{J}(W^r), W^{r+1} - W^r \rangle \\
&+ \frac{M}{2}\|W^{r+1} - W^r\|^2
\end{split}
\end{equation}

Let's define the expected local model update as:
\begin{equation}
\Delta_r = \sum_{i} \pi_i (w_i^{r+1} - W^r)
\end{equation}

Then:
\begin{equation}
W^{r+1} = W^r + \Delta_r
\end{equation}

Substituting:
\begin{equation}
\mathcal{J}(W^{r+1}) \leq \mathcal{J}(W^r) + \langle \nabla \mathcal{J}(W^r), \Delta_r \rangle + \frac{M}{2}\|\Delta_r\|^2
\end{equation}

From Lemma 1, we know the local models make progress. Using Assumption \ref{asmp:het} (bounded client heterogeneity):

\begin{equation}
\begin{split}
    \langle \nabla \mathcal{J}(W^r), \Delta_r \rangle \leq -\eta S(1 - M\eta)\|\nabla \mathcal{J}(W^r)\|^2 \\+ \eta SG\|\nabla \mathcal{J}(W^r)\| + \gamma\|\Delta_r\|
\end{split}
\end{equation}

And $\|\Delta_r\|$ can be bounded by $\eta S(\|\nabla \mathcal{J}(W^r)\| + G + \gamma)$.

Substituting and taking expectation:
\begin{equation}
    \begin{split}
        \mathbb{E}[\mathcal{J}(W^{r+1})] \leq \mathbb{E}[\mathcal{J}(W^r)] + \frac{M\eta^2 S^2}{2}(\sigma^2 + G^2 + \gamma^2) \\
+ \eta SG\|\nabla \mathcal{J}(W^r)\|  - \eta S(1 - M\eta - \frac{M\eta S}{2})\|\nabla \mathcal{J}(W^r)\|^2
    \end{split}
\end{equation}

For small enough $\eta$, we can simplify this to:
\begin{equation}
\begin{split}
    \mathbb{E}[\mathcal{J}(W^{r+1})] \leq \mathbb{E}[\mathcal{J}(W^r)]  + \frac{M\eta^2 S}{2}(\sigma^2 + G^2 + \gamma^2)\\ + \eta SG\|\nabla \mathcal{J}(W^r)\|  - \frac{\eta S}{4}(1 - M\eta)\|\nabla \mathcal{J}(W^r)\|^2
\end{split}
\end{equation}

Rearranging:
\begin{equation}
\begin{split}
    \mathbb{E}[\mathcal{J}(W^r) - \mathcal{J}(W^{r+1})] \geq \frac{\eta S}{4}(1 - M\eta)\|\nabla \mathcal{J}(W^r)\|^2 \\
    - \eta SG\|\nabla \mathcal{J}(W^r)\| - \frac{M\eta^2 S}{2}(\sigma^2 + G^2 + \gamma^2)
\end{split}
\end{equation}
\end{proof}

\begin{table*}[ht!]
\centering
\small
\begin{tabular}{l ccccccc}
\toprule
Method & OrganCMNIST & OrganSMNIST & PathMNIST & FMNIST & CIFAR-10 & CIFAR-100 \\
\midrule
FedAvg \cite{mcmahan2017communication}    &  36.66  &  30.89  &  48.07   &  61.20  &  50.88  &  40.11  \\
\rowcolor{gray!15}
\quad + FLFA               &  \textbf{40.02}  &  \textbf{33.69}  &  \textbf{54.03}  &  \textbf{61.71}  &  \textbf{51.74}  &  \textbf{42.42}  \\
\midrule
FedProx \cite{FedProx}     &  41.29  &  \textbf{34.93}  &  49.00  &  72.05  &  49.18  &  38.88  \\
\rowcolor{gray!15}
\quad + FLFA               &  \textbf{43.38}  &  33.54  &  \textbf{53.21}  &  \textbf{72.85}  &  \textbf{50.79}  &  \textbf{41.10}  \\
\midrule
MOON  \cite{MOON}      &  40.32  &  \textbf{30.60}  &  42.94  &  56.44  &  47.50  & 40.05  \\
\rowcolor{gray!15}
\quad + FLFA               &  \textbf{41.54}  &  29.96  &  \textbf{49.12}  &  \textbf{62.81}  &  \textbf{49.52}  & \textbf{45.12} \\
\midrule
FedAvgM \cite{FedAvgM}     &  51.65 &  42.52  &  58.97  &  70.92  &  53.10  &  41.53  \\
\rowcolor{gray!15}
\quad + FLFA               &  \textbf{52.22}  &  \textbf{42.73}  &  \textbf{55.34}  &  \textbf{71.11}  &  \textbf{54.94}  &  \textbf{41.83} \\
\midrule
FedRS \cite{li2021fedrs}     &  56.04  &  38.13  &  68.48  &  74.39  &  \textbf{59.00}  &  41.45 \\
\rowcolor{gray!15}
\quad + FLFA               &  \textbf{57.00}  &  \textbf{44.22}  &  \textbf{65.60}  &  \textbf{76.47}  &  58.29  & \textbf{41.60}  \\
\midrule
FedLC \cite{fedlc}        &  55.75  &  42.66  &  62.99 &  73.93  &  58.46  &   40.13 \\
\rowcolor{gray!15}
\quad + FLFA               &  \textbf{56.40}  &  \textbf{42.82}  &  \textbf{66.61}  &  \textbf{75.69}  &  \textbf{59.51}  &  \textbf{40.72}  \\
\midrule
\rowcolor{gray!15}
\textbf{Avg. gain}         &  \textbf{+1.48}  &  \textbf{+1.21}  &  \textbf{+2.24}  &  \textbf{+1.95}  &  \textbf{+1.11}  &  \textbf{+1.77}  \\
\bottomrule
\end{tabular}
\caption{Test accuracy (\%) comparison on various FL algorithms with and without FLFA \textbf{under extreme data heterogeneity ($\mathbf{\beta=0.1}$)}. FLFA shows remarkable robustness in this challenging setting, consistently outperforming baselines with an average gain of up to \textbf{+2.24\%p}.}
\label{tab:accuracy_b01}
\end{table*}

\section{Additional Experiment Settings}
\label{sec:sup_exp_settings}
The source code for reproducing our experimental results is provided in the attached supplementary zip file.

\textbf{Training details.} Across all experiments, we use Stochastic Gradient Descent (SGD) as the optimizer with a momentum of 0.9, an initial learning rate of 0.01, and weight decay of 0.001. The batch size for local training is set to 64. We apply a learning rate decay of 0.998 after each communication round. The final test accuracy reported in our main paper is the average of the accuracies from the last 10\% of the total communication rounds.

\subsection{Dataset Details}
We evaluated our proposed method on a diverse set of benchmarks, including medical image datasets from MedMNIST \cite{medmnist} and standard computer vision datasets. The details of the datasets used in our experiments are as follows.

\begin{itemize} 
\item \textbf{PathMNIST}: A dataset based on colon pathology, consisting of 9 classes. It contains 89,996 training samples and 7,180 test samples. 
\item \textbf{BloodMNIST}: A dataset of blood cell microscope images classified into 8 classes. It comprises 11,959 training samples and 3,421 test samples. 
\item \textbf{OrganCMNIST}: A dataset consisting of abdominal CT scans (Coronal view) with 11 classes. It provides 12,975 training samples and 8,216 test samples. 
\item \textbf{OrganSMNIST}: Similar to OrganCMNIST but based on the Sagittal view of abdominal CT scans, containing 11 classes. It includes 13,932 training samples and 8,827 test samples. 
\end{itemize}

In addition to medical datasets, we employed widely used vision datasets to assess generalizability. 
\begin{itemize} 
\item \textbf{Fashion-MNIST (FMNIST)}: A dataset of grayscale clothing images with 10 classes, consisting of 60,000 training samples and 10,000 test samples. 
\item \textbf{CIFAR-10 and CIFAR-100}: Large-scale datasets containing color images classified into 10 and 100 classes, respectively. Both datasets consist of 50,000 training samples and 10,000 test samples. 
\item \textbf{ImageNet-100}: A subset of the ImageNet dataset containing 100 classes. It consists of approximately 130,000 training samples and 5,000 test samples. 
\end{itemize}

\textbf{Non-IID data partitioning.} To simulate realistic FL scenarios with heterogeneous data, we adopt a non-IID data partitioning scheme using a Dirichlet distribution. We set the concentration parameter $\beta = 0.3$, where a smaller $\beta$ value indicates a more heterogeneous data distribution across clients. For $N$ total clients and each class $c \in \{1,\ldots,C\}$, the data allocation for each client is determined by sampling a proportion vector $\rho_c \in \mathbb{R}^N$ for each class $c$ as follows:
\begin{equation}
    \rho_c \sim \text{Dir}(\beta\mathbf{1}_N)
\end{equation}
where $\mathbf{1}_N$ is an $N$-dimensional vector of ones. This process is repeated for all classes to create a non-IID distribution of data among the clients.

\textbf{Hyperparameters.} The hyperparameters for the experiments are detailed in source code with experimental settings.

\textbf{Computing Resources.} Unless otherwise specified, experiments were conducted on NVIDIA RTX 4000 Ada GPU. The experiments involving $\beta$ and client scalability utilized RTX 8000 GPU, while ResNet-50 and participation rate experiments were performed on NVIDIA V100 GPU.

\begin{table*}[ht!]
\centering
\small
\begin{tabular}{l ccccccc}
\toprule
Method & BloodMNIST & OrganCMNIST & OrganSMNIST & PathMNIST & FMNIST & CIFAR-10 & CIFAR-100 \\
\midrule
FedAvg \cite{mcmahan2017communication}   &  48.81  &  67.65  &  53.26  &  66.21  & 81.02  &  78.37 &  51.78 \\
\rowcolor{gray!15}
\quad + FLFA             &  \textbf{62.00}  &  \textbf{72.23}  &  \textbf{56.58}  &  66.21 &  \textbf{81.20}  & \textbf{79.16}  &  \textbf{52.83} \\
\midrule
FedProx \cite{FedProx}   & \textbf{60.99} & 71.04  &  51.00  &  67.89 &  \textbf{80.80} &  75.52 &  \textbf{52.96} \\
\rowcolor{gray!15}
\quad + FLFA             & 59.05  &  \textbf{71.05}  &  \textbf{58.42}  &  \textbf{68.03} &  80.74  & \textbf{76.52}  & 52.45  \\
\midrule
MOON  \cite{MOON}     &  49.14  &  \textbf{71.40}  &  54.86  & 63.90  & \textbf{81.03}  & 78.49  &  52.87 \\
\rowcolor{gray!15}
\quad + FLFA             &  \textbf{66.46}  &  69.06  &  \textbf{56.45}  &  \textbf{67.03} &  80.07  & \textbf{79.08}  & \textbf{53.47}  \\
\midrule
FedAvgM \cite{FedAvgM}   &  73.17  &  74.34  &  \textbf{62.01} & 66.16  &  87.03  & 77.29  & \textbf{53.21}  \\
\rowcolor{gray!15}
\quad + FLFA             &  \textbf{77.67}  &  \textbf{74.46}  &  61.87  & \textbf{73.34}  &  \textbf{87.11} & \textbf{80.25}  & 53.02  \\
\midrule
FedRS \cite{li2021fedrs}   & 80.16  &  74.17  &  \textbf{60.22}  & 72.22  &  84.78  &  80.65 & \textbf{52.53}  \\
\rowcolor{gray!15}
\quad + FLFA             & \textbf{80.76}  &  \textbf{75.62}  &  59.64  & \textbf{73.55}  &  \textbf{85.58} &  \textbf{80.93} &  51.98 \\
\midrule
FedLC \cite{fedlc}      &  81.09 &  75.95 &  60.01  &  68.39 & 84.59  & 74.82  &  \textbf{51.21} \\
\rowcolor{gray!15}
\quad + FLFA             &  \textbf{82.19}  &  \textbf{76.29}  &  \textbf{61.41}  &  \textbf{71.36} & \textbf{86.11}  & \textbf{81.15}  & 50.88 \\
\midrule
\rowcolor{gray!15}
\textbf{Avg. gain}       &  \textbf{+5.80}  &  \textbf{+0.69}  &  \textbf{+2.17}  &  \textbf{+2.46}  &  \textbf{+0.26}  &  \textbf{+1.99}  &  \textbf{0.01}  \\
\bottomrule
\end{tabular}
\caption{Test accuracy (\%) comparison on various FL algorithms with and without FLFA \textbf{using the ResNet-50 architecture}. FLFA demonstrates strong generalizability to deeper models, achieving significant performance improvements with an average gain of up to \textbf{+5.80\%p}.}
\label{tab:accuracy_resnet50}
\end{table*}

\begin{table*}[ht!]
\centering
\small
\begin{tabular}{l cccccc}
\toprule
Method & OrganCMNIST & OrganSMNIST & PathMNIST & FMNIST & CIFAR-10 & CIFAR-100 \\
\midrule
FedAvg \cite{mcmahan2017communication}   &  71.92 & 52.29  & 66.26  &  79.77 & 58.10  & \textbf{36.29}   \\
\rowcolor{gray!15}
\quad + FLFA             &  \textbf{73.69} & \textbf{53.19}  & \textbf{67.70}  & \textbf{81.01} &  \textbf{59.46} &  36.14  \\
\midrule
FedProx \cite{FedProx}    &  71.74 &  52.20 &  \textbf{68.90}  &  81.12 & 56.50  & 36.01    \\
\rowcolor{gray!15}
\quad + FLFA             &  \textbf{73.54} &  \textbf{52.82} &  68.07   & \textbf{82.69}  & \textbf{58.10}  & \textbf{36.84}   \\
\midrule
MOON  \cite{MOON}      &  71.71  & 53.95 & 67.25  & 80.95  & 56.22  &  \textbf{36.16}  \\
\rowcolor{gray!15}
\quad + FLFA             & \textbf{72.96}  & \textbf{54.05}  & \textbf{68.47}  & \textbf{81.89} &  \textbf{58.75} &  35.98  \\
\midrule
FedAvgM \cite{FedAvgM}    & 72.69  &  \textbf{58.48} & 71.92  & 83.53 &  \textbf{59.75} &  35.75   \\
\rowcolor{gray!15}
\quad + FLFA              & \textbf{76.95}  &  57.39 & \textbf{72.56}  & \textbf{83.65} &  59.47 & \textbf{36.27}   \\
\midrule
FedRS \cite{li2021fedrs}    &  75.17 &  55.96 &  75.09  & 83.66 & 58.77  & \textbf{36.69}   \\
\rowcolor{gray!15}
\quad + FLFA              &  \textbf{76.93} &\textbf{ 57.29 } & 75.09  &  \textbf{84.06} & \textbf{59.48}  & 36.59   \\
\midrule
FedLC \cite{fedlc}       & 73.14  & 57.75  &  72.90  & \textbf{84.72} & 58.07  & 35.61   \\
\rowcolor{gray!15}
\quad + FLFA              &  \textbf{74.36} & \textbf{57.95}  &  \textbf{73.29}  &  84.35  & \textbf{59.40}  & \textbf{36.67}   \\
\midrule
\rowcolor{gray!15}
Avg. gain              &  \textbf{+2.01} &\textbf{ +0.34} &  \textbf{+0.48}  &  \textbf{+0.65}  & \textbf{+1.21}  & \textbf{+0.33}   \\
\bottomrule
\end{tabular}
\caption{Test accuracy (\%) comparison on various FL algorithms with and without FLFA \textbf{when the number of clients is scaled up to 200}. FLFA consistently outperforms baselines across all datasets, achieving an average gain of up to \textbf{+2.01\%p}, demonstrating its scalability in large-scale networks.}
\label{tab:accuracy_clients}
\end{table*}

\section{Experiment metrics} \label{sec:supp:metrics}
To evaluate the performance and behavior of our method, we use the following metrics. For the feature-based metrics, let $\mathbf{f}_i \in \mathbb{R}^d$ be the $d$-dimensional feature representation of a sample $i$ from the model's penultimate layer, and let $y_i \in \{1, \dots, C\}$ be its corresponding class label.

\textbf{Intra-class variance.} This metric measures the average compactness of the feature representations within each class. A smaller value indicates that features for the same class are more tightly clustered. It is calculated by finding the average Euclidean distance from each sample to its corresponding class centroid (mean feature vectors). Let $\mathcal{D}_c$ be the set of samples belonging to class $c$, and let the class centroid be $\boldsymbol{\mu}_c = \frac{1}{|\mathcal{D}_c|} \sum_{i \in \mathcal{D}_c} \mathbf{f}_i$. The intra-class variance is then:
\begin{equation}
    d_{\text{intra}} = \frac{1}{C} \sum_{c=1}^{C} \left( \frac{1}{|\mathcal{D}_c|} \sum_{i \in \mathcal{D}_c} \|\mathbf{f}_i - \boldsymbol{\mu}_c\|_2 \right)
\end{equation}

\textbf{Inter-class variance.} This metric measures the average separation between different classes in the feature space. A larger value indicates that the representations of different classes are more distinct. It is calculated as the average Euclidean distance between the centroids of all unique pairs of classes. The inter-class variance is then:
\begin{equation}
    d_{\text{inter}} = \frac{2}{C(C-1)} \sum_{c=1}^{C-1} \sum_{c'=c+1}^{C} \|\boldsymbol{\mu}_c - \boldsymbol{\mu}_{c'}\|_2
\end{equation}

\textbf{Separability ratio.} This is a holistic measure of feature quality, defined as the ratio of the inter-class variance to the intra-class variance. A higher separability ratio signifies a more discriminative feature space, where classes are both well-separated from each other and internally compact.
\begin{equation}
    \text{Separability Ratio} = \frac{d_{\text{inter}}}{d_{\text{intra}}}
\end{equation}

\textbf{Local drift.} This metric quantifies the divergence among client updates within a single communication round. Let $\Delta \mathbf{w}_i$ be the concatenated and flattened weight update vector from client $i$ across all model layers. The local drift is calculated as the average Euclidean distance between each client's model update and the mean update, $\Delta \bar{\mathbf{w}} = \frac{1}{K} \sum_{i=1}^{K} \Delta \mathbf{w}_i$:
\begin{equation}
    \mathcal{H} = \frac{1}{K} \sum_{i=1}^{K} \|\Delta \mathbf{w}_i - \Delta \bar{\mathbf{w}}\|_2
\end{equation}
The local drift reduction by FLFA is calculated as:
\begin{equation}
    \mathcal{H}_{BP} - \mathcal{H}_{FLFA}
\end{equation}
A positive value from this calculation signifies that FLFA was more effective at reducing client drift than the standard BP baseline in that round.

\begin{table*}[ht!]
\centering
\small
\begin{tabular}{l ccccccc}
\toprule
Method & BloodMNIST & OrganCMNIST & OrganSMNIST & PathMNIST & FMNIST & CIFAR-10 & CIFAR-100 \\
\midrule
FedAvg \cite{mcmahan2017communication}   & \textbf{70.26}  &  \textbf{80.88} &  \textbf{67.25} &  \textbf{64.75} & 81.54  &  72.01 &  54.27 \\
\rowcolor{gray!15}
\quad + FLFA             &  69.18 & 80.64 & 67.09  &  64.21 & \textbf{82.00}  &  \textbf{74.70} & \textbf{54.41}  \\
\midrule
FedProx \cite{FedProx}   &  71.63 &  \textbf{82.19} &  66.88 &  78.71 &  88.90 & 73.86  &  54.01 \\
\rowcolor{gray!15}
\quad + FLFA             &  \textbf{73.37} &  81.36 & \textbf{66.90}  &  \textbf{82.55} & \textbf{89.54}  & \textbf{75.09}  &  \textbf{54.13} \\
\midrule
MOON  \cite{MOON}     &  \textbf{70.55} & 80.88  & 65.34  &  67.08 & 82.80  & 72.78  &  54.52 \\
\rowcolor{gray!15}
\quad + FLFA             & 70.49  & \textbf{81.22}  & \textbf{66.24}  & \textbf{67.60}  &  \textbf{80.38} & \textbf{75.44}  &  \textbf{54.75} \\
\midrule
FedAvgM \cite{FedAvgM}   & 78.71  & 81.72  &  66.84 & 63.87 &  68.40 &  78.66 &  54.59 \\
\rowcolor{gray!15}
\quad + FLFA             & \textbf{81.85}  & \textbf{82.28}  &  \textbf{68.07} & \textbf{65.92}  &  85.71 &  \textbf{80.94} & \textbf{56.29}  \\
\midrule
FedRS \cite{li2021fedrs}   &  83.76 & 81.87  & \textbf{67.98}  &  74.77 & 87.56  &  78.14 &  53.98 \\
\rowcolor{gray!15}
\quad + FLFA             &  \textbf{82.71} &  \textbf{82.95} & 67.84  &  \textbf{75.08} &  \textbf{88.16} & \textbf{78.82}  &  \textbf{56.22} \\
\midrule
FedLC \cite{fedlc}      &  83.12 & 81.47  &  \textbf{68.79} &  77.85 & 86.11  & 79.69  &  54.74 \\
\rowcolor{gray!15}
\quad + FLFA             &  \textbf{85.13} &  \textbf{82.46} & 68.07  &  \textbf{78.86} &  \textbf{88.26} &\textbf{80.94}   &  \textbf{54.76} \\
\midrule
\rowcolor{gray!15}
Avg. gain              &  \textbf{+0.78} & \textbf{+0.32}  &  \textbf{+0.19}  &  \textbf{+1.20}  & \textbf{+3.12}  & \textbf{+1.80}  & \textbf{+0.74}   \\
\bottomrule
\end{tabular}
\caption{Test accuracy (\%) comparison on various FL algorithms with and without FLFA \textbf{when trained with extended local epochs (15)}. Despite the increased risk of client drift, FLFA effectively anchors updates, resulting in significant performance improvements with an average gain of up to \textbf{+3.12\%p}.}
\label{tab:accuracy_epochs}
\end{table*}

\begin{table*}[ht!]
\centering
\small
\begin{tabular}{l ccccccc}
\toprule
Method & BloodMNIST & OrganCMNIST & OrganSMNIST & PathMNIST & FMNIST & CIFAR-10 & CIFAR-100 \\
\midrule
FedAvg \cite{mcmahan2017communication}   &  64.66  &  68.63 &  \textbf{46.50} &  54.82 &  73.86 &  61.05  &  43.49 \\
\rowcolor{gray!15}
\quad + FLFA             &  \textbf{68.79} & \textbf{68.88}  &  44.81  & \textbf{55.13}  & \textbf{74.64}  & \textbf{61.21}  & \textbf{43.55}  \\
\midrule
FedProx \cite{FedProx}   &  63.43  &  70.30 &  45.91 &  59.15  &  73.37 &  62.19  &  43.50  \\
\rowcolor{gray!15}
\quad + FLFA             & \textbf{64.59}  &  \textbf{71.08} &  \textbf{47.19} & \textbf{60.31}  &  \textbf{74.32} &  \textbf{63.02} &  \textbf{44.56}  \\
\midrule
MOON  \cite{MOON}     &  60.99 &  \textbf{70.12}  & 43.52  &  57.59 &  \textbf{74.62} &  61.31 &  44.33 \\
\rowcolor{gray!15}
\quad + FLFA             &  \textbf{62.21}  &  68.77  & \textbf{45.43}  & \textbf{58.79} & 74.60  & \textbf{61.61}  &  \textbf{45.75} \\
\midrule
FedAvgM \cite{FedAvgM}   & 70.68  &  72.27 & \textbf{48.54}  &  \textbf{70.44}  & 77.94  & 64.46  & 43.48  \\
\rowcolor{gray!15}
\quad + FLFA             & \textbf{71.13}  &  \textbf{73.08} &  47.97  &  67.00 &  \textbf{78.24} & \textbf{64.77} &  \textbf{43.59} \\
\midrule
FedRS \cite{li2021fedrs}   &  \textbf{74.32} &  75.76  &  52.41 &  71.77 & 81.45  &  67.14  &  \textbf{43.88} \\
\rowcolor{gray!15}
\quad + FLFA             &  74.00 & \textbf{76.01}  &  \textbf{54.07}  & \textbf{75.51}  &  \textbf{81.87}  & \textbf{68.58}  &  43.49  \\
\midrule
FedLC \cite{fedlc}      &  75.04 &  \textbf{75.38} &  \textbf{54.67}  & \textbf{72.58}  & \textbf{82.31} & 66.82  & 42.89  \\
\rowcolor{gray!15}
\quad + FLFA             & \textbf{75.79}  &  74.76 & 54.55  &  71.47 & 81.61  & \textbf{67.66}  &  \textbf{43.20}  \\
\midrule
\rowcolor{gray!15}
\textbf{Avg. gain}       &  \textbf{+1.23}  &  \textbf{+0.02}  &  \textbf{+0.41}  &  \textbf{+0.31}  &  \textbf{+0.29}  &  \textbf{+0.65}  &  \textbf{+0.43}  \\
\bottomrule
\end{tabular}
\caption{Test accuracy (\%) comparison on various FL algorithms with and without FLFA \textbf{when the participation rate is 0.05}. Even with a low participation rate, FLFA maintains training stability and achieves positive average gains across all scenarios (up to \textbf{+1.23\%p}).}
\label{tab:accuracy_smp005}
\end{table*}

\begin{table*}[ht!]
\centering
\small
\begin{tabular}{l ccc ccc ccc}
\toprule
\multirow{2}{*}{Method} & \multicolumn{3}{c}{BloodMNIST} & \multicolumn{3}{c}{OrganCMNIST} & \multicolumn{3}{c}{OrganSMNIST} \\
\cmidrule(lr){2-4}\cmidrule(lr){5-7}\cmidrule(lr){8-10}
\# of FA layers & 1 & 2 & 3 & 1 & 2 & 3 & 1 & 2 & 3 \\
\midrule
FedAvg \cite{mcmahan2017communication}   & \textbf{63.11} & 60.76 & 58.80 & \textbf{73.78} & \textbf{73.78} & 71.35 & 55.94 & 56.20 & \textbf{57.65} \\
FedProx \cite{FedProx}   & \textbf{61.04} & 58.58 & 59.85 & \textbf{73.86} & 73.31 & 72.80 & 55.94 & \textbf{56.23} & 54.89 \\
MOON  \cite{MOON}        & 61.02 & \textbf{61.77} & 59.75 & 72.62 & 73.86 & \textbf{75.71} & 56.33 & \textbf{56.93} & 56.05 \\
FedAvgM \cite{FedAvgM}   & \textbf{66.42} & 65.96 & 63.09 & 76.23 & \textbf{77.26} & 76.41 & 59.39 & 58.67 & \textbf{60.24} \\
FedSOL \cite{lee2024fedsol} & \textbf{61.79} & 61.17 & 61.72 & 73.24 & \textbf{74.29} & 74.27 & 58.07 & 57.48 & \textbf{58.56} \\
FedRS \cite{li2021fedrs} & \textbf{76.86} & 76.52 & 75.38 & 74.96 & 75.03 & \textbf{75.32} & \textbf{58.13} & 56.39 & 57.09 \\
FedLC \cite{fedlc}       & \textbf{76.27} & 74.18 & 73.67 & 75.63 & 74.95 & \textbf{76.52} & \textbf{58.14} & 58.01 & 58.03 \\
FedACG \cite{fedacg}     & 62.13 & 63.50 & \textbf{65.36} & \textbf{74.87} & 73.45 & 73.87 & 54.68 & 55.54 & \textbf{56.16} \\
\midrule
\rowcolor{gray!15}
\textbf{Avg. gain}       & \textbf{--} & \textbf{-0.78} & \textbf{-1.38} & \textbf{--} & \textbf{+0.09} & \textbf{+0.13} & \textbf{--} & \textbf{-0.15} & \textbf{+0.26} \\
\bottomrule
\end{tabular}
\caption{Test accuracy (\%) comparison with varying numbers of FA layers (Part 1). Avg. gain is computed relative to 1 FA layer. \textbf{Note that all results in this table are obtained using a single fixed seed for fair comparison across FA layers}; thus, the values for 1 layer may differ from the main results. Best results are in \textbf{bold}.}
\label{tab:accuracy_fa_1}
\end{table*}

\begin{table*}[ht!]
\centering
\small
\begin{tabular}{l ccc ccc ccc}
\toprule
\multirow{2}{*}{Method} & \multicolumn{3}{c}{PathMNIST} & \multicolumn{3}{c}{FMNIST} & \multicolumn{3}{c}{CIFAR-10} \\
\cmidrule(lr){2-4}\cmidrule(lr){5-7}\cmidrule(lr){8-10}
\# of FA layers & 1 & 2 & 3 & 1 & 2 & 3 & 1 & 2 & 3 \\
\midrule
FedAvg \cite{mcmahan2017communication}   & \textbf{69.34} & 69.02 & 67.05 & \textbf{81.71} & 81.49 & 80.05 & \textbf{66.36} & 64.27 & 65.30 \\
FedProx \cite{FedProx}   & \textbf{67.08} & 64.51 & 69.86 & \textbf{80.22} & 79.75 & 78.56 & 64.56 & \textbf{65.78} & 64.22 \\
MOON  \cite{MOON}        & \textbf{72.79} & 69.45 & 68.34 & \textbf{80.94} & 80.31 & 78.61 & 66.15 & \textbf{67.86} & 64.01 \\
FedAvgM \cite{FedAvgM}   & 76.75 & \textbf{77.58} & 74.40 & \textbf{85.81} & 84.99 & 85.36 & 64.76 & 65.70 & \textbf{66.39} \\
FedSOL \cite{lee2024fedsol} & 67.34 & \textbf{69.30} & 67.85 & 80.56 & \textbf{80.71} & 79.91 & 67.80 & 67.65 & \textbf{69.76} \\
FedRS \cite{li2021fedrs} & \textbf{76.54} & 74.40 & 74.70 & \textbf{86.30} & 86.18 & 86.10 & 68.90 & 67.04 & \textbf{71.48} \\
FedLC \cite{fedlc}       & \textbf{74.88} & 72.22 & 73.12 & \textbf{86.36} & 85.91 & 85.63 & 68.76 & \textbf{70.50} & 68.39 \\
FedACG \cite{fedacg}     & \textbf{70.55} & 64.51 & 69.86 & 81.37 & \textbf{82.28} & 81.17 & 63.97 & 66.06 & \textbf{66.55} \\
\midrule
\rowcolor{gray!15}
\textbf{Avg. gain}       & \textbf{--} & \textbf{-1.79} & \textbf{-1.26} & \textbf{--} & \textbf{-0.21} & \textbf{-0.99} & \textbf{--} & \textbf{+0.45} & \textbf{+0.61} \\
\bottomrule
\end{tabular}
\caption{Test accuracy (\%) comparison with varying numbers of FA layers (Part 2). Avg. gain is computed relative to 1 FA layer. \textbf{Note that all results in this table are obtained using a single fixed seed for fair comparison across FA layers}; thus, the values for 1 layer may differ from the main results. Best results are in \textbf{bold}.}
\label{tab:accuracy_fa_2}
\end{table*}

\begin{figure*}[ht!]
    \centering
    \includegraphics[width=0.8\linewidth]{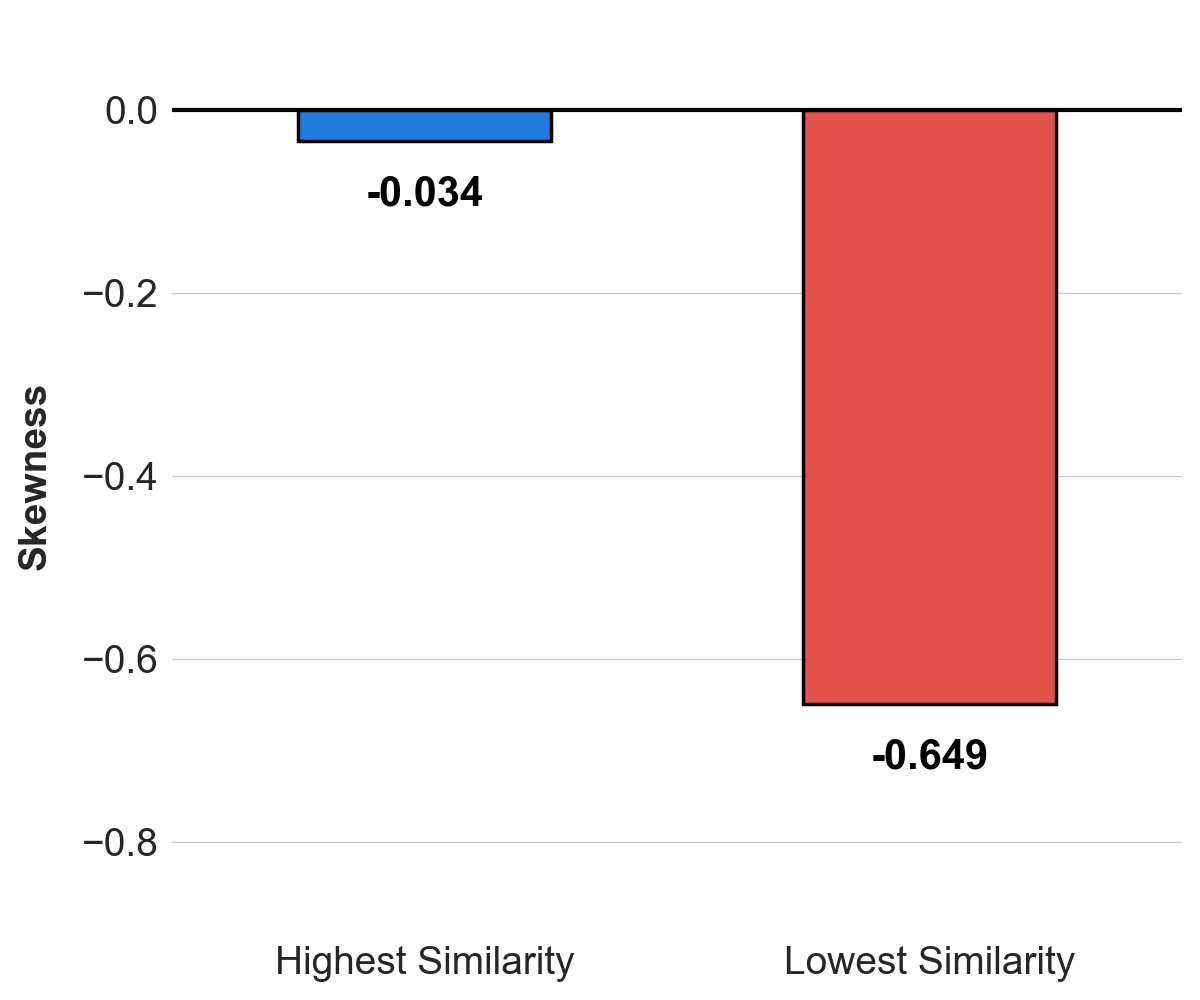}
    \caption{\textbf{Distribution of cosine similarities between global and local gradients at Round 0.} The distribution is \textbf{skewed towards lower values}, indicating that most layers suffer from gradient divergence. Based on this skewness, we adopt the 'Lowest' selection strategy to target the layer with the most severe misalignment (minimum similarity) for applying FLFA.}
    \label{fig:flfa_skewness}
\end{figure*}

\section{Additional Robustness Experiments} \label{sec:supp:add_exp}
To validate the robustness of our proposed method, we conducted addtional experiments under various challenging conditions. We utilized diverse datasets including MedMNIST and CIFAR, and applied FLFA across multiple FL algorithms. Note that BloodMNIST was excluded from the scalability (increasing clients) and extreme heterogeneity experiments. Due to its limited total sample size, partitioning it under these rigorous settings would result in an insufficient number of samples per client, making meaningful local training infeasible.

\subsection{Extreme Data Heterogeneity}
We simulated an extremely heterogeneous environment by setting the Dirichlet parameter $\beta=0.1$. It is worth noting that values below 0.1 typically result in training failure due to excessive data skew; thus, this setting represents a near-maximum limit of heterogeneity. Under this harsh condition, as shown in Table~\ref{tab:accuracy_b01}, FLFA exhibited remarkable robustness, securing substantial Avg. gains of +1.11\% to +2.24\%. This confirms that FLFA is particularly effective when data distributions are highly non-IID.

\subsection{Complex Model Architecture}
To assess applicability across architectures, we applied FLFA to ResNet-50 \cite{he2016deep}, which is significantly deeper than MobileNetV2. Deeper models are generally more prone to optimization difficulties in FL due to the larger parameter space. The results in Table~\ref{tab:accuracy_resnet50} indicate that FLFA retains its performance advantage on the ResNet-50 architecture, achieving a maximum Avg. gain of +5.80\%, thereby validating its compatibility with complex models.

\subsection{Scalability with More Clients}
Increasing the total number of clients significantly reduces the frequency of each client's participation with smaller data for each client, making global convergence more difficult and unstable. We evaluated the scalability of FLFA by increasing the number of clients to 200. As presented in Table~\ref{tab:accuracy_clients}, FLFA consistently outperforms the baselines across all datasets and algorithms. Notably, our method achieved an Avg. gain ranging from +0.33\% to +2.01\%.

\subsection{Impact of Extended Local Epochs}
Training with a large number of local epochs (set to 15) poses a risk of severe client drift, as local models tend to diverge significantly from the global model. Table~\ref{tab:accuracy_epochs} shows that FLFA effectively mitigates this issue by anchoring local updates. Despite the increased local training epochs, FLFA maintained superior performance with an Avg. gain between +0.19\% and +3.12\%, proving its capability to leverage extended local epochs.

\subsection{Low Participation Rate}
We further tested the method in a scenario with a low sample fraction of 0.05 (Table~\ref{tab:accuracy_smp005}). A lower participation rate increases the variance of the aggregated global model, leading to unstable convergence. Even in this setting, FLFA demonstrated stability, recording positive Avg. gains across all cases (up to +1.23\%), which suggests that our feedback mechanism helps maintain training stability despite sparse client participation.

\subsection{Evaluation on Additional Baselines} 
We originally conducted comparative experiments with SCAFFOLD and FedDyn. However, these results were omitted from the main text and presented here because both algorithms failed to converge on certain datasets under our experimental settings. Specifically, SCAFFOLD showed instability across multiple tasks, and FedDyn failed to converge on PathMNIST and FMNIST. Integrating FLFA into FedDyn provided consistent performance improvements as shown in Table~\ref{tab:feddyn}.

\begin{table*}[ht!]
\centering
\small
\begin{tabular}{l cccccc}
\toprule
Method & BloodMNIST & OrganCMNIST & OrganSMNIST & PathMNIST & FMNIST & CIFAR-10 \\
\midrule
FedDyn \cite{acar2021federated}   & 71.95  &  74.51  &  61.24 & -  &  - &  74.81 \\
\rowcolor{gray!15}
\quad + FLFA             &  \textbf{73.91} &  \textbf{75.73} & \textbf{62.01}  & -  &  - &  \textbf{74.83} \\
\bottomrule
\end{tabular}
\caption{Test accuracy (\%) comparison on FedDyn with and without FLFA. FLFA successfully enhances the performance of FedDyn.}
\label{tab:feddyn}
\end{table*}

\section{Additional Ablation Studies} \label{sec:supp:add_abl}
We further investigate the impact of the number of FLFA layers and validate the effectiveness of our layer selection strategy based on the distribution of gradient cosine similarities.

\subsection{Number of FA Layers}
We analyzed the effect of increasing the number of layers applying FLFA ($N_{FA}$) from 1 to 3. To ensure a strictly fair comparison across different $N_{FA}$ settings, all results in this ablation study (Table~\ref{tab:accuracy_fa_1} and Table~\ref{tab:accuracy_fa_2}) are reported based on a single fixed seed. Consequently, the reported accuracies for $N_{FA}=1$ differ slightly from the main results, which are averaged over three random seeds. As shown in Table~\ref{tab:accuracy_fa_2}, for complex datasets such as CIFAR-10, increasing $N_{FA}$ to 2 or 3 yields marginal performance gains (Avg. gain of +0.61\%). This indicates that applying feedback alignment to multiple layers can provide additional stability in handling complex feature representations. However, the results also confirm that the default setting of $N_{FA}=1$ is sufficiently robust, achieving competitive performance with minimal computational overhead. For simpler datasets, $N_{FA}=1$ generally outperforms or matches multi-layer configurations, suggesting that a single anchor point is adequate for rectifying local drift in most scenarios.

\subsection{Layer Selection Strategy via Skewness} 
Our layer selection strategy is guided by the distribution of cosine similarities between the global and local gradients, measured at the initial round. Figure~\ref{fig:flfa_skewness} illustrates this distribution. We observe the skewness of these similarity scores to determine the optimal intervention strategy. 

As shown in Figure~\ref{fig:flfa_skewness}, the distribution is often skewed towards the lower end, indicating that most layers suffer from gradient divergence. In such cases, we employ the \textbf{'Lowest'} strategy, selecting the layer with the minimum similarity to aggressively correct the most severe bottleneck of gradient divergence. Conversely, if not, the 'Highest' strategy would be viable to reinforce existing alignment. 

\subsection{Selected layers}
We tracked the layer selection frequency throughout the training process using MobileNetV2. Table~\ref{tab:layer_selection_ratio} summarizes the selection ratios for each stage under the 'Lowest' and 'Highest' strategies.

The results reveal a distinct contrast in behavior. The \textbf{'Lowest'} strategy predominantly selects shallow layers, with the Stem and Stages 1--3 accounting for over 70\% of the selections. This empirical evidence suggests that gradient divergence---caused by data heterogeneity---is most severe in the early feature extraction layers, prompting our algorithm to prioritize rectifying these initial blocks. Conversely, the \textbf{'Highest'} strategy shifts its focus toward deeper layers (e.g., Stage 4, Stage 6) and the final classifier. This implies that high-level semantic features tend to maintain relatively higher gradient consistency across clients compared to low-level features.

Consequently, this distribution confirms that the 'Lowest' strategy effectively targets the most vulnerable components of the model (early stages) to mitigate local drift, aligning with our hypothesis that correcting the point of maximum divergence is crucial for FL performance.

\begin{table}[t]
    \centering
    \resizebox{0.95\columnwidth}{!}{%
    \begin{tabular}{lccc}
        \toprule
        \textbf{Model Block} & \textbf{Stage} & \textbf{Lowest (\%)} & \textbf{Highest (\%)} \\
        \midrule
        Stem & Pre-processing & 17.69 & 16.37 \\
        \midrule
        \multirow{7}{*}{Bottleneck} 
        & Stage 1 & \textbf{24.96} & 9.76 \\
        & Stage 2 & 15.54 & 3.91 \\
        & Stage 3 & 13.84 & 9.63 \\
        & Stage 4 & 12.47 & \textbf{20.92} \\
        & Stage 5 & 3.75 & 0.05 \\
        & Stage 6 & 1.33 & \textbf{20.47} \\
        & Stage 7 & 0.33 & 1.67 \\
        \midrule
        Head & Conv1 & 0.07 & 1.97 \\
        Classifier & Final Output & 10.02 & 15.25 \\
        \bottomrule
    \end{tabular}%
    }
    \caption{Distribution of selected layers in MobileNetV2 under different selection strategies. The 'Lowest' strategy prioritizes early stages where gradient divergence is severe, whereas the 'Highest' strategy frequently selects deeper stages.}
    \label{tab:layer_selection_ratio}
\end{table}

\end{document}